\documentclass[balance,upint,subscriptcorrection,varvw,mathalfa=cal=boondoxo,american,greek,pdf-a,fontspec,colorlinks]{asmeconf}
\usepackage{float}
\usepackage{units}
\usepackage{tikz}
\usepackage{enumitem}
\usepackage[none]{hyphenat}
\setlist[enumerate]{itemsep=0mm}  
\usepackage{hyperref}
\usepackage[capitalise, nameinlink, noabbrev]{cleveref}
\creflabelformat{equation}{#2\textup{#1}#3}  
\hypersetup{%
	pdfauthor={Thomas Nakken Larsen},
	pdftitle={Low-dimensional latent encoding of a high-dimensional range-finding sensor for reinforcement learning based collision avoidance},                  
	pdfkeywords={Variational Autoencoder, Autonomous Surface Vessel, Reinforcement Learning, Collision Avoidance},
}

\begin{document}
	

	\title{Variational Autoencoders for exteroceptive perception in reinforcement learning-based collision avoidance}
	
	\SetAuthors{%
    Thomas Nakken Larsen\affil{1},
        Eirik Runde Barlaug\affil{1},
    Adil Rasheed\affil{1}\CorrespondingAuthor{adil.rasheed@ntnu.no}
}

\SetAffiliation{1}{Department of Engineering Cybernetics, NTNU, O.S. Bragstadsplass 2, 7034, Trondheim, Norway}
	
	
	\maketitle
	\begingroup
	\renewcommand
	\thefootnote{\textsection}
	\endgroup
	
	
	
	\normalfont
	

\begin{abstract}
Modern control systems are increasingly turning to machine learning algorithms to augment their performance and adaptability. Within this context, Deep Reinforcement Learning (DRL) has emerged as a promising control framework, particularly in the domain of marine transportation. Its potential for autonomous marine applications lies in its ability to seamlessly combine path-following and collision avoidance with an arbitrary number of obstacles. However, current DRL algorithms require disproportionally large computational resources to find near-optimal policies compared to the posed control problem when the searchable parameter space becomes large. To combat this, our work delves into the application of Variational AutoEncoders (VAEs) to acquire a generalized, low-dimensional latent encoding of a high-fidelity range-finding sensor, which serves as the exteroceptive input to a DRL agent. The agent's performance, encompassing path-following and collision avoidance, is systematically tested and evaluated within a stochastic simulation environment, presenting a comprehensive exploration of our proposed approach in maritime control systems.
\end{abstract}

	
	
\section{Introduction}
Modern control systems are increasingly turning to machine learning algorithms to augment their performance and adaptability. Within this context, Deep Reinforcement Learning (DRL) has emerged as a promising control framework, particularly in the domain of marine transportation. Its potential lies in its ability to seamlessly combine path-following and collision avoidance, ushering in a new era of intelligent, autonomous agents. The deployment of autonomous agents in maritime settings is not merely a technological endeavor but one with significant safety and efficiency implications. These intelligent agents have the capacity to optimize fuel consumption and, perhaps more crucially, substantially reduce the frequency of human-caused accidents in this sector. 

Achieving autonomy through DRL hinges on two essential components. First, it leverages guidance-theoretic features to enhance navigation capabilities, enabling precise control. Second, it relies on high-fidelity exteroceptive sensing as the foundation for the autonomous agent's obstacle perception, allowing it to navigate challenging environments effectively. While traditional control methods may manage collision-free navigation when dealing with a limited number of known obstacles, DRL agents equipped with these capabilities can adapt to address an array of previously unknown obstacles dynamically. However, the key to this capability lies in the meaningful encoding of high-fidelity sensor data to enable collision-free navigation.

To address this issue, some studies suggest using VAE-based approaches to learn useful latent representations of data for downstream reinforcement learning tasks \cite{intro_yarats2020improving, intro_andersen2018dreaming}, especially for pixel data.
However, a thorough search of the literature reveals that despite the widespread application of VAEs in various contexts, their integration into the realm of DRL for maritime control and collision avoidance remains relatively unexplored. This research gap motivates the exploration of VAE-based feature extraction to further improve upon the works of \cite{general_hansen_risk-based_2023}, with the aim of extracting more useful representations of the environment for the downstream DRL task. The agent's performance, encompassing path-following and collision avoidance, is systematically tested and evaluated within a stochastic simulation environment, presenting a comprehensive exploration of this innovative approach in maritime control systems. To guide this study, the following set of research questions are explored:

\setlist{nolistsep}
\begin{itemize}[noitemsep]
    \item Can the VAE feature extractor be successfully integrated into a DRL agent within the simulated maritime environment?  
    \item How do model complexity and hyperparameters influence the VAE's data reproduction and generalization capabilities?
    \item What is the performance difference between DRL agents equipped with pre-trained VAE-encoders of different complexity and how do they compare to a non-VAE DRL agent?
\end{itemize}
The rest of the paper consists of theory, methodology, results and discussions, and conclusion sections. The theory section is kept short and relevant literuares are cited where needed.

\section{Preliminaries} \label{sec:Theory}
\subsection{Vessel dynamics} \label{sec:Theory:Dynamics}
\begin{figure}[h!]
    \centering
    \includegraphics[width=\linewidth]{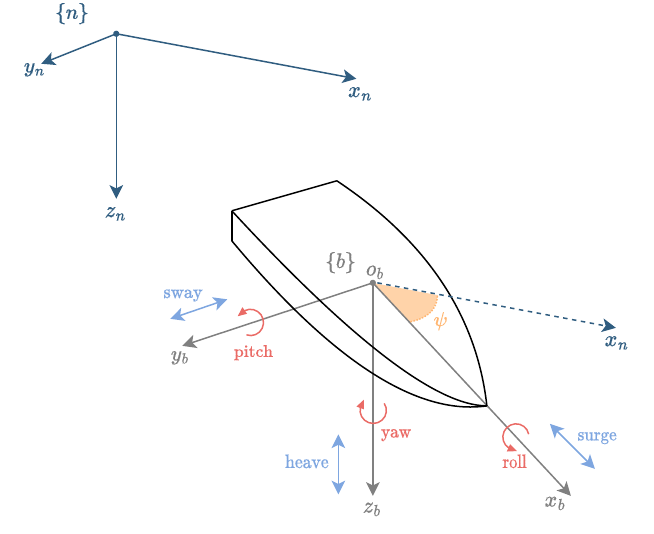}
    \caption{Illustration of the inertial and body-fixed reference frames relevant for vessel dynamics.}
    \label{fig:frames_of_reference}
\end{figure}

Frames of reference must be defined to model the ship dynamics accurately. Here, two conventions are used: The \textit{North-East-Down} (NED) reference frame and the \textit{body-fixed} reference frame. The NED frame $\{n\} = (x_n, y_n, z_n)$ is defined as the tangent plane to the Earth's surface with its axes oriented towards true \textit{north}, true \textit{east} and \textit{down} (towards the Earth's center) denoted $x_n, y_n, z_n$, respectively. Assuming the ship operates within a localized area with nearly constant longitude and latitude, this constant Earth-fixed tangent plane, denoted $\{n\}$, is used. The body-fixed reference frame, denoted as ${b} = (x_b, y_b, z_b)$, is attached to the vessel. The $x_b$-axis runs from the rear to the front, the $y_b$-axis extends towards the right side (starboard), and the $z_b$-axis goes from the top of the ship downwards towards the earth's center. The origin $o_b$ is positioned at the midpoint of the ship, at the waterline level. These conventions are illustrated in \cref{fig:frames_of_reference}.

The state of a ship is commonly presented as a 6 degrees-of-freedom (DOF) system, encompassing three translational modes: surge, sway, and heave, and three rotational modes: roll, pitch, and yaw. Each mode corresponds to movement along or rotation about one of the three axes $(x_b, y_b, z_b)$, respectively. Using the standardized SNAME notation \cite{theory_society1950nomenclature}, the system variables are defined as
\begin{equation}\label{eq:6DOF system vars}
\begin{aligned}
    \boldsymbol{\eta} &= \begin{bmatrix} x_n & y_n & z_n & \phi & \theta & \psi \end{bmatrix}^{T}, \\
    \boldsymbol{\nu} &= \begin{bmatrix}u &  v & w & p & q & r\end{bmatrix}^T.
\end{aligned}
\end{equation}
Here, $\boldsymbol{\eta}$ denotes the state vector comprising the generalized coordinates. The first three entries represent the vessel's position relative to $\{n\}$, while the latter three (roll, pitch, yaw) describe the orientation of $\{b\}$ with respect to $\{n\}$ following a $\text{zyx}$ Euler angle convention. Note that $\psi$ is commonly referred to as the heading angle. The entries of $\boldsymbol{\nu}$ are surge, sway, heave, roll rate, pitch rate, and yaw rate, respectively. 

Two fundamental assumptions from \cite{theory_fossen_modeling_2004} are adopted: 1) The vessel is always located at the sea surface ($z_n \approx 0$), exhibiting no heave, pitch or roll motion ($w = \phi = p = \theta = q \approx 0$) and 2) The vessel is unaffected by external disturbances like waves, currents, or wind. Under assumption 1 the state variables of the system reduce to
\begin{equation}\label{eq:3DOF system vars}
    \boldsymbol{\eta} = \begin{bmatrix}x_n & y_n & \psi\end{bmatrix}^T, \quad \boldsymbol{\nu} = \begin{bmatrix}u & v & r\end{bmatrix}^T.
\end{equation}

The ship dynamics can then be expressed in 3-DOF under assumptions 1 and 2 as
\begin{align}\label{eq: 3-DOF dynamics}
    \boldsymbol{\dot{\eta}} & = \textbf{R}_{z,\psi}(\boldsymbol{\eta})\boldsymbol{\nu} \\
    \textbf{M}\boldsymbol{\dot{\nu}} + \textbf{C}(\boldsymbol{\nu})\boldsymbol{\nu} + \textbf{D}(\boldsymbol{\nu})\boldsymbol{\nu} & = \textbf{B}\boldsymbol{f},
\end{align}
where $\textbf{R}_{z,\psi} \in SO(3)$ describes the simple rotation about the z-axis between the Earth-fixed velocity vector $\boldsymbol{\dot{\eta}}$ and the body-fixed velocity vector $\boldsymbol{\nu}$, as expressed by
\begin{equation}\label{eq: rotation matrix R_z_psi}
\textbf{R}_{z,\psi} = \begin{bmatrix}
        \cos\psi & -\sin\psi & 0 \\
        \sin\psi & \cos\psi & 0 \\
        0 & 0 & 1
    \end{bmatrix}.
\end{equation}
Moreover, the mass matrix $\textbf{M} \in \mathbb{R}^{3\times3}$ integrates both the rigid-body mass and added mass. The Coriolis matrix $\textbf{C}(\boldsymbol{\nu}) \in \mathbb{R}^{3\times3}$ accounts for centripetal and Coriolis forces. The damping effects are represented by the matrix $\textbf{D}(\boldsymbol{\nu}) \in \mathbb{R}^{3\times3}$, and $\textbf{B} \in \mathbb{R}^{3\times2}$ is the actuator configuration matrix. Lastly, $\boldsymbol{f} = \begin{bmatrix} T_u & T_r\end{bmatrix}^T$ is the control vector consisting of the surge force and the yaw moment, respectively. A third entry for the bow thruster force is omitted due to its ineffectiveness at high speeds, as demonstrated in \cite{theory_sorensen_ship_2017}.
    
\subsection{Marine guidance}
\begin{figure}
    \centering
    \includegraphics[width=\linewidth]{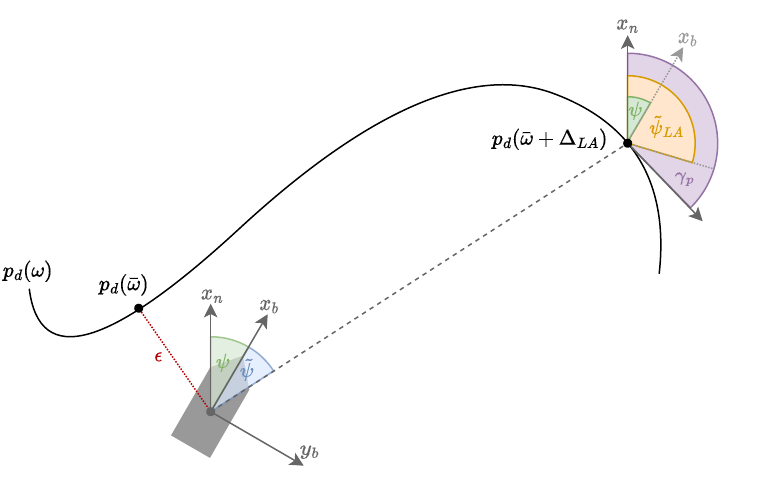}
    \caption{Graphical representation of the guidance features relevant for path-following.}
    \label{fig:path}
\end{figure}

This subsection provides an overview of the path-following theory for curved trajectories, which is essential when we introduce the navigational observation vector later. The desired path (parameterized by the continuous path-variable $\omega$) is represented by $\boldsymbol{p}_d(\omega) = [x_d(\omega),y_d(\omega)]^T$, where $x_d$ and $y_d$ are given in NED-frame. Then $\Bar{\omega}$ is let to denote the value of the path variable that minimizes the Euclidean distance between the vessel's position and the desired path. This minimal distance, namely the \textit{cross-track error} (CTE), is thus given by
\begin{equation}\label{eq:CTE}
    \epsilon = \Big|\Big| [x_n,y_n]^T - \boldsymbol{p}_d(\Bar{\omega}) \Big|\Big|.
\end{equation}
Furthermore, the look-ahead point $\boldsymbol{p}_d(\Bar{\omega} + \Delta_{LA})$ is the point that lies a constant (user-defined) look-ahead distance $\Delta_{LA}$ further along the path from $\boldsymbol{p}_d(\Bar{\omega})$. Using this,  \textit{heading error} $\Tilde{\psi}$ may be defined as 
\begin{equation}\label{eq:heading_error}
    \Tilde{\psi} = \text{atan2}\left(\frac{y_d(\Bar{\omega} + \Delta_{LA}) - y_n}{x_d(\Bar{\omega} + \Delta_{LA}) - x_n}\right) - \psi,
\end{equation}
which is the heading correction needed to steer directly toward the look-ahead point from the current vessel position $[x_n,y_n]^T$, with the current heading $\psi$. The \textit{look-ahead heading error}, denoted as $\Tilde{\psi}_{LA}$, is identified as the angle discrepancy between the direction of the path at the look-ahead point and the current heading of the vessel. Specifically, it is expressed as:
\begin{equation}\label{eq:look_ahead_heading_error}
    \Tilde{\psi}_{LA} = \gamma_p(\Bar{\omega} + \Delta_{LA}) - \psi,
\end{equation}
where $\gamma_p(\omega)$ is the path-angle, defined as the difference between north ($x_n$) and the tangent of the path at $\omega$. \cref{fig:path} provides a graphical representation of the introduced guidance features.

    \subsection{Deep reinforcement learning} \label{sec:Theory:DRL}
    Deep reinforcement learning (DRL) is a subfield of machine learning that incorporates deep learning techniques into reinforcement learning algorithms. Reinforcement learning, at its core, involves training an agent to make sequential decisions by interacting with an environment and adapting to feedback in the form of scalar rewards or penalties. The learning algorithm is tasked with finding the set of parameters that maximize the agent's expected return. What sets DRL apart is the integration of artificial neural networks for parameterizing the agent, enabling the acting policy to learn complex representations and patterns. This approach has proven particularly successful in autonomous decision-making tasks where traditional methods struggle, such as playing games \cite{silver2016mastering,Badia2020Agent57OT} and robotic control \cite{gu2017drlRoboticManipulation,yunlong2023OCvsDRLquadrotors}. DRL algorithms, like Deep Q-Networks (DQN) \cite{theory_mnih_playing_2013_DQN} and Proximal Policy Optimization (PPO) \cite{theory_schulman_PPO_2017}, have demonstrated remarkable achievements in mastering complex tasks by learning from trial and error. 
    
    \subsection{Variational autoencoders} \label{sec:Theory:VAE}
    The Variational Autoencoder (VAE) \cite{kingma2013vae} is a generative model that combines the principles of autoencoders with variational inference techniques. In a VAE, the encoder network maps input data to a set of variational latent distributions, which comprise a probabilistic representation of the underlying structure in the data. This set of distributions is parameterized by both means and variances, allowing stochasticity in the encoding process. The encoder, therefore, not only encodes the input data but also captures its likelihood in the latent space. During training, VAEs aim to minimize the Kullback-Leibler (KL) divergence between the variational latent distribution and a predefined prior distribution. This KL-divergence minimization acts as a regularizer, encouraging the model to learn a well-structured and continuous latent space. The decoder, on the other hand, reconstructs data from samples in this latent space. \Cref{eq:vae_loss} describes the VAE loss function, containing both reconstruction and KL-divergence terms. A common extension to the standard VAE formulation is the $\beta$-VAE \cite{higgins2017bVAE}, which introduces a parameter $\beta$ that scales the KL-term for enhanced regularization and latent variable disentanglement. This variant is adopted in this work. By combining the encoding and decoding processes with the variational inference framework, VAEs excel in tasks such as image generation \cite{razavi2019imgGenVAE}, image compression \cite{minnen2018VAEimgCompression}, and representation learning \cite{vandenoord2017VAEreprLearning}, providing a powerful tool for capturing complex data distributions.
    \begin{equation}\label{eq:vae_loss}
    \begin{aligned}
        \mathcal{L}_{VAE}(\boldsymbol{x}, \boldsymbol{\phi}, \boldsymbol{\theta}) = & - \sum_{i=1}^N \mathbb{E}_{\boldsymbol{z}_i \sim q_{\phi}(\boldsymbol{z}_i|\boldsymbol{x}_i)}[\log p_{\boldsymbol{\theta}}(\boldsymbol{x}_i|\boldsymbol{z}_i)] \\
        & - D_{KL}(q_{\boldsymbol{\phi}}(\boldsymbol{z}_i|\boldsymbol{x}_i) || p_{\boldsymbol{\theta}}(\boldsymbol{z}_i)),
    \end{aligned}
    \end{equation}
    \vspace{-7mm}
\section{Methodology} \label{sec:Method}
    The approach introduced in this work aims to employ a VAE to obtain a generalized, low-dimensional latent encoding of a high-fidelity range-finding sensor. This latent encoding will serve as the exteroceptive input to a DRL agent tasked with guiding an autonomous surface vessel (ASV). That is, the encoder component of the VAE will, post-training, be integrated directly as a feature extractor into the DRL agent. Consequently, this method has a dual purpose: firstly, it explores the capacity of the VAE to extract meaningful information from the DRL agent's environment, and secondly, it investigates the effectiveness of the encoder when employed as a feature extractor for the DRL agent. 

    \begin{figure}[h!]
        \centering
        \includegraphics[width=\linewidth]{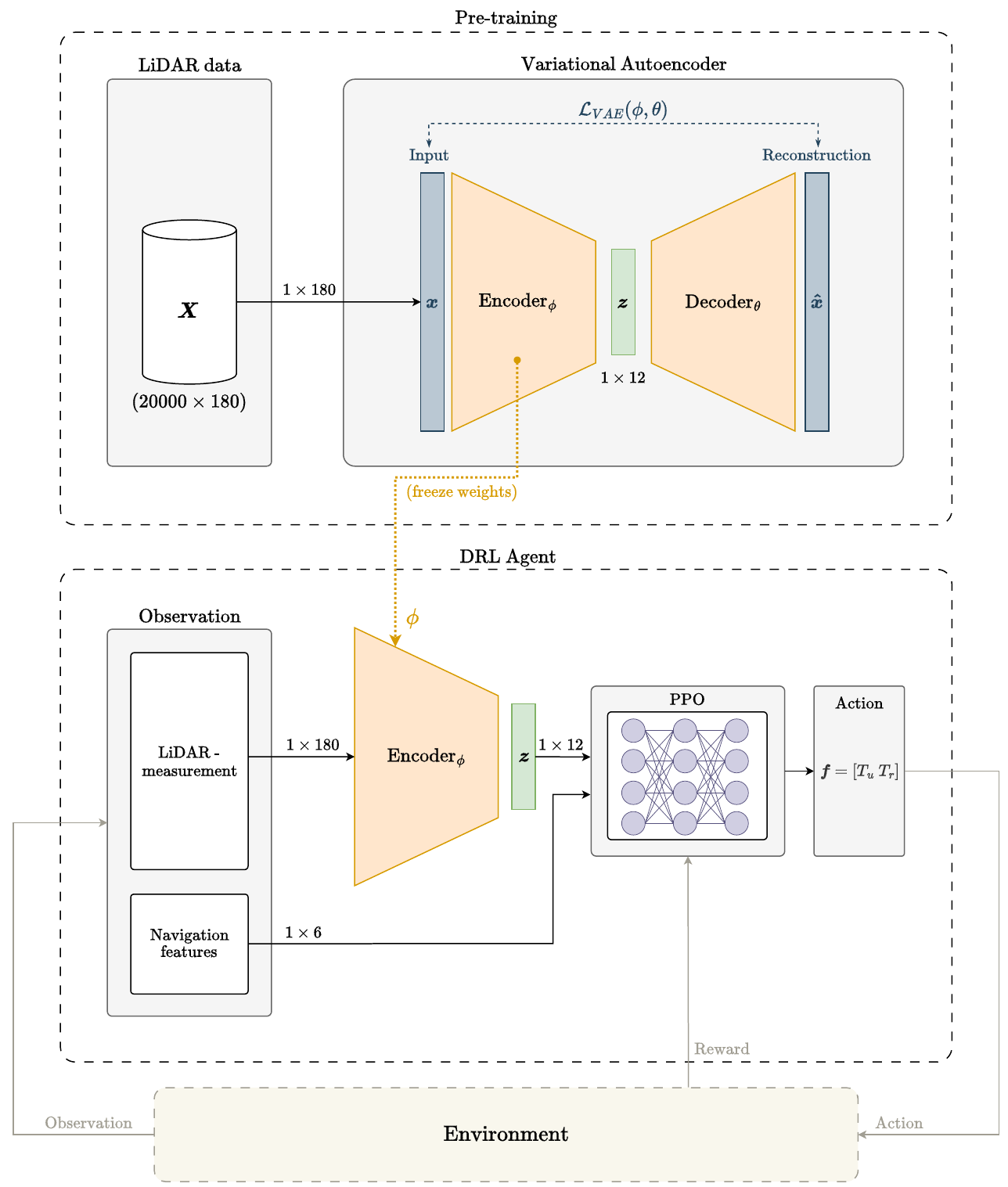}
        \caption{High-level illustration of the proposed VAE+DRL system architecture and training protocols.}
        \label{fig:architecture_highlevel}
    \end{figure}
    
    A high-level diagram of the full architecture of this work is depicted in \cref{fig:architecture_highlevel}. In this figure, the top view describes the pre-training of the VAE. After training, the encoder parameters are copied to the DRL agent and implemented as a feature extractor. The latent encoding of the LiDAR measurement, $\boldsymbol{z}$, the navigational features of the agent, and a reward from the environment are fed into the PPO algorithm, which outputs an action to the environment.

    This section is organized into three parts. First, an overview of the simulation framework and DRL training environment to which our work serves as an extension is provided. Subsequently the details of the VAE approach are presented. Finally, the performance metrics of both the VAE and the DRL agent are presented.
    
    \subsection{DRL for path following and collision avoidance}
    The DRL agents are trained and tested in the autonomous vessel software framework developed by \citep{general_meyer_et_al_2020_taming} and later extended by \citep{general_larsen_comparing_2021,vaaler2023modular}, simulating a Cybership II \citep{theory_fossen_modeling_2004} surface vessel under the calm sea assumption. This miniature replica of a supply ship is tasked to follow a randomly generated path while avoiding static obstacles (landmasses) and dynamic obstacles (other vessels). The episode ends when any of the following events has occurred: 1) The vessel is within $5m$ from the end of the path, 2) the vessel's progress along the path has exceeded $99\%$, 3) the vessel has collided, 4) the number of time steps has exceeded $2000$ or 5) the cumulative reward is lower than $-2000$. Several model-free DRL algorithms have been previously evaluated for this control problem, among which PPO is the most reliable algorithm \citep{general_larsen_comparing_2021}, thus motivating our choice of DRL algorithm in this work.
    
    \subsection{State and action spaces}
\begin{figure}[h]
        \centering
        \includegraphics[width=\linewidth]{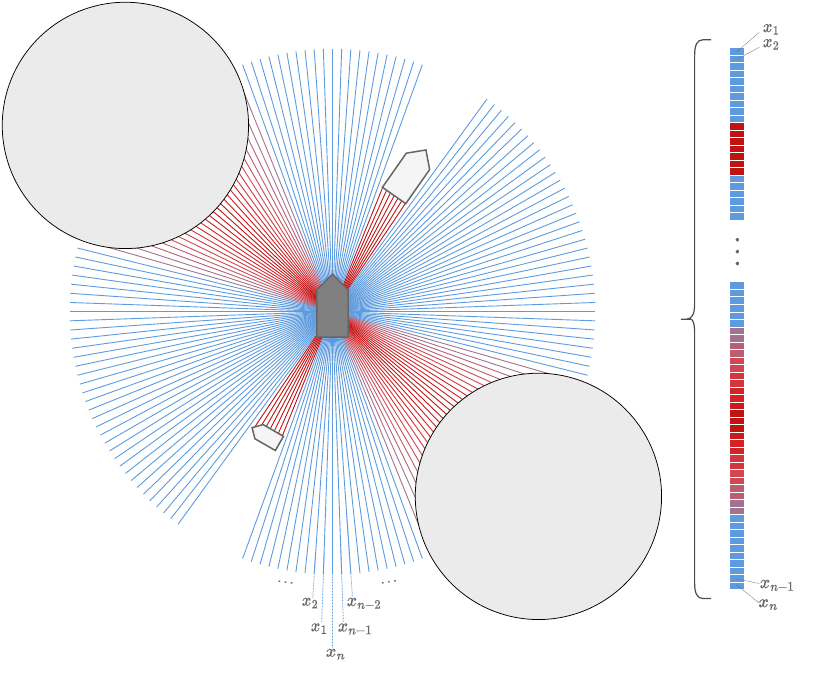}
        \caption{Range-finding sensor suite and its mapping from circular sensor data to a perception vector.}
        \label{fig:rangefinding_sensor}
    \end{figure}

    A state and action space for the agent to facilitate autonomous navigation within the DRL framework is defined. The state space comprises an observation vector consisting of \textit{navigation features} and \textit{perception features}. The navigation features, given by 
    \begin{equation}\label{eq:nav_features}
        I_{nav} = \begin{bmatrix}u& v& r& \epsilon& \Tilde{\psi}& \Tilde{\psi}_{LA}\end{bmatrix}^T
    \end{equation}
    describes the navigational state of the ASV relative to the desired path. The simulation framework is equipped with a planar high-fidelity range-finding sensor suite to enable exteroception for the DRL agent, allowing it to sense its surroundings. The sensor suite provides LiDAR-like measurements describing distances to surrounding objects relative to the body frame in a 360-degree span around the vessel. The sensors have a maximum measurable distance of $150m$ and are evenly spaced around the vessel with a sensor every two degrees, resulting in $180$ sensors. \Cref{fig:rangefinding_sensor} shows a top-down view of the sensor suite and its mapping to the perception vector. The dark gray vessel at the center represents the ownship, while the light gray objects indicate static (circular) or dynamic (polygon) obstacles. Ray-colorations from blue to red indicate the closeness to the obstacles. Note that the first and last elements of the perception vector are neighboring elements in reality.

    The navigation features are stacked with a latent encoding of the perception vector before being fed into the agent. The action space is defined by the two-dimensional control input vector $\boldsymbol{f} = [T_u\ T_r]^T$ as discussed in \cref{sec:Theory:Dynamics}.
    
    \subsection{Reward function}
    The design of the reward function is critical for guiding the agent towards desired policies. A well-crafted reward function steers the agent away from undesirable states and promotes efficient and accurate path following. In our simulated environment, with large episodic time horizons and many potential modes of failure per episode, a sparse reward signal is assumed to be infeasible. 
    
    The reward function used in this work closely resembles the one obtained in \cite{general_larsen_comparing_2021}. The reward in timestep $(t)$ is given by 
    \begin{equation}\label{eq:reward}
        r^{(t)} = \begin{cases} r_{\text{collision}}, &  \text{if collision} \\
        r^{(t)}_{\text{path}} - r_{\text{exists},} & \text{otherwise}, \end{cases}
    \end{equation}
    where $r_{\text{collision}} = -1000$ incentivizes collision avoidance and $r_{\text{exists}} = 1$ penalizes the agent for existing (i.e., inducing a sense of urgency to complete the episode). The final term of \cref{eq:reward}, namely $r_{\text{path}}^{(t)}$, consists of three terms itself and is given by 
    \begin{equation}\label{eq:r_path}
        r_{\text{path}}^{(t)} = \underbrace{\frac{u^{(t)}}{u_{\text{max}}}}_{\text{Speed term}} \cdot\ \underbrace{(1 + \text{cos}(\Tilde{\psi}^{(t)}))}_{\text{Heading term}}\ \cdot\ \underbrace{\frac{1}{1+|\epsilon^{(t)}|}}_{\text{CTE term}}.
    \end{equation}

    \subsection{VAE-based range sensor encoding}
    Directly coupling the range sensor to the input of a fully connected network for DRL is unlikely to result in a sufficient policy for path following and collision avoidance \cite{general_meyer_et_al_2020_taming}. Instead, the agent should be provided with a low-dimensional representation of its environment to increase the probability of the DRL algorithm finding a good policy. Obtaining this representation, aiming to be as informative and meaningful as possible, is the primary focus of this section. To achieve this, the specifics of the VAE setup necessary for replicating the study is delved into.

    \subsubsection{Data generation and augmentation}
    This study utilizes a dataset comprising 10,000 distinct range observations obtained through manual navigation within the simulated environment. Dynamic obstacles are initialized with random heading angles and velocities in the interval $[0.1, 0.2]\text{ms}^{-1}$, while static circular obstacles are placed randomly along the path. Note that since these are placed \textit{along the path}, only a subset of them appear within the maximum measurable distance of the vessel at a time.

    To increase the volume and improve the balance of the data, synthetic observations are generated according to the methodology outlined in \cite{Meyer2020}. This process yields another 10,000 samples, grouped as follows:
    \begin{itemize}
    \setlist[itemize]{noitemsep}
        \item 5,000 of the observations mirrored the original dataset, featuring a mix of three static and two dynamic obstacles randomly placed within the $x_{max}\times x_{max}$-grid surrounding the vessel. The locations of these obstacles are uniformly distributed, with dynamic obstacles also having uniformly distributed heading angles within $[0,2\pi]\text{rad}$. The sizes of the obstacles follow two Poisson distributions with means of 25 and 10, respectively.
        \item 2,500 of the observations comprises scenarios with five dynamic obstacles exclusively.
        \item 2,500 of the observations comprises scenarios with five static obstacles exclusively.
    \end{itemize}
    
    The 20,000 by 180-dimensional dataset, denoted as $\boldsymbol{X}$, is further augmented by applying arbitrary rotations of the data along the second axis (i.e., about the z-axis of the vessel), expanding the number of observations to 60,000. This transformation is informed by preliminary experiments indicating that such rotations potentially provide rotational invariance to the models.
    
    Furthermore, the distance measurements are normalized and flipped by applying the transformation $x_k = 1 - \tfrac{x_k}{x_{max}}$ such that $x_k \in [0,1]\ \forall\ k$, where $x_{max} = 150m$ is the maximum measurable distance. In this context, any measurement $x_k=1$ corresponds to a collision, and $x_k=0$ corresponds to an absence of obstacles within the detectable range of that sensor.
    
    The final dataset is partitioned into training and test sets with a 70:30 split. A subset comprising 20\% of the training set is used for validation. Finally, to enhance robustness, a small Gaussian noise ($\mathcal{N}(0,\sigma^2)$, where $\sigma^2 = 0.007$) is added to the training data, with $\sigma$ being empirically determined through visual inspection of the data and noise. This noise corresponds to augmenting the original data with a standard deviation of approx $12.55m$.

    \subsubsection{VAE architecture}
    
    As indicated by the findings in \cite{general_hansen_risk-based_2023}, both the shallow (1-layer) and 3-layer CNN configurations demonstrated promising outcomes when implemented as feature extractors for the DRL agent. Motivated by these results, these architectures for the encoders in our VAEs and their transposed versions for the decoders were adopted. The convolutional blocks given in \cref{tab:conv_blocks} are investigated, with model names given by the depth of the encoder and decoder.
    \begin{table}[h]
        \centering
        \setlength{\tabcolsep}{12pt}
        \caption{Parameters of the encoders and decoders of the two tested VAE configurations.}
        \begin{tabular}{@{}lll@{}}
        \toprule
         \textbf{Parameter} & \textbf{ShallowConvVAE} & \textbf{DeepConvVAE} \\
         \hline
         No. layers & 1 & 3 \\
         Channels & (1) & (3, 2, 1) \\
         Kernel sizes & (45) & (45, 3, 3) \\
         Strides & (15) & (15, 1, 1) \\
         Paddings & (15) & (15, 1, 1) \\
         Padding modes & `circular' (x1) & `circular' (x3) \\
         \bottomrule
        \end{tabular}
        \label{tab:conv_blocks}
    \end{table}
    In all cases, the decoders are the transposes of the encoders, with the same values as given in the table. Following the convolutional block, a fully connected layer maps the representations to the latent variables, and at the end of the decoder, the Sigmoid activation function is applied.
    
    \subsubsection{Circularly padded transposed convolution}
    The adoption of circular padding in both the encoder and decoder of the VAE addresses the inherent circularity of the perception vector, where the first and last elements are adjacent in the simulation. Standard zero-padding, which treats these elements as unrelated, introduces discontinuities and edge artifacts that may disrupt the learning process, leading to suboptimal model performance.
    
    Preliminary experiments showed that using zero-padding in the decoder led to poor reconstructions, especially at the junction of the leading and trailing elements of the perception vector. This observation was crucial in guiding our methodology. Although zero-padding in transposed convolution is trivial to implement and is a standard feature in deep learning libraries such as PyTorch \cite{Paszke2019PyTorch} or Keras \cite{general_chollet2015keras}, circular padding for transposed convolutional layers is less common and not natively supported. To address this issue, developed a one-dimensional circularly padded transposed convolution method, which we will refer to as \textsc{CPTC-1D} was developed. This approach extends the concepts presented in \cite{general_schubert_circular_2019}, adapting them to a one-dimensional context suitable for our application within a PyTorch framework.

    \subsection{Experimental setup and performance evaluation}
    The PPO implementation from Stable-Baselines3 \cite{raffin2021sb3} is used to train the agent. The chosen hyperparameters for the algorithm are obtained from \cite{general_larsen_comparing_2021}, reproduced in \cref{tab:ppo_params}, as these gave good results in their study.
    
    \begin{table}
        \centering
        \caption{Hyperparameters for the PPO algorithm. The non-default values are highlighted in bold.}
        \small
        \begin{tabular}{@{}llr@{}}
        \toprule
        \textbf{Hyperparameter} & \textbf{Description} & \textbf{Value} \\
             \hline
            \verb|learning_rate| & Learning rate & \textbf{2e-4} \\
            \verb|n_steps| & Steps to run per env. per update  & \textbf{1024} \\
            \verb|batch_size| & Minibatch size  & \textbf{32} \\
            \verb|n_epochs| & Epochs for updating surrogate loss  & \textbf{4} \\
            \verb|gamma| & Discount factor & \textbf{0.999} \\
            \verb|gae_lambda| & Bias vs. variance factor for GAE & \textbf{0.98} \\
            \verb|clip_range| & Clipping & 0.2 \\
            \verb|clip_range_vf| & Clipping for the value function & None \\
            \verb|normalize_advantage| & Normalize the advantage, or not & True \\
            \verb|ent_coef| & Entropy coeff. & \textbf{0.01} \\
            \verb|vf_coef| & Value function coeff. & 0.5 \\
            \verb|max_grad_norm| & Max gradient clipping & 0.5 \\
            \bottomrule
        \end{tabular}
        \label{tab:ppo_params}
    \end{table}

    \subsubsection{VAE evaluation} \label{sec:Method:VAE_eval}
    Both models are implemented in Python using PyTorch and trained on approximately 42,000 samples with a learning rate of $0.001$ and a batch size of $64$ for $25$ epochs using the Adam optimizer \cite{kingma2014adam}. The loss function is based on \cref{eq:vae_loss}, using binary cross-entropy and $\beta$-scaled KL divergence. Models are trained across $10$ random seeds and shuffles of the data, reporting the average training and validation errors with associated $95\%$ confidence intervals (CIs). The size of the latent space bottleneck is set to 12, matching the output size of the convolutional filter; multiple sizes (1, 2, 6, 12, and 24) were preliminarily investigated, and it was found that matching the output size from the last convolutional layer yielded the best results.
    
    The estimated distribution of the latent representation is visualized using both a shallow and a deep VAE across $\beta \in \{0, 1, 5\}$ to inspect the regularization effects on the latent space for our specific data.
    
    The models are tested on approximately 18,000 observations from the test set to obtain an empirical estimate of the generalization error. Furthermore, samples from the test set are used to create reconstructions of the input vectors, plotted with polar coordinates to appear identical to the simulation. Randomly drawn test samples are also used to visualize the latent distributions of a given input along with its reconstruction.

    \subsubsection{DRL evaluation}\label{sec:performance_metrics_drl}
    The DRL agent was evaluated by comparing the VAE+PPO model against a baseline and several VAE+PPO configurations against each other. The baseline architecture is the best-performing CNN-perception feature extractor from \cite{general_hansen_risk-based_2023}, namely a 1-layer CNN with a similar architecture as our ShallowConvVAE-encoder. Instead of directly adopting the results from \cite{general_hansen_risk-based_2023}, the findings from the paper were independently replicated within the current framework. For the comparison of different VAE+PPO models, the following feature extractor configurations were tested:
    \begin{enumerate} 
        \item ShallowConvVAE with \textit{locked parameters}, meaning the PPO algorithm is not allowed to update the weights of the feature extractor during training. 
        \item ShallowConvVAE with \textit{unlocked parameters}; the agent is allowed to update the weights of the feature extractor during DRL training. 
        \item DeepConvVAE with locked parameters.
        \item DeepConvVAE with unlocked parameters.
        \item \textbf{Baseline}: A CNN with equivalent architecture as the encoder of ShallowConvVAE. Randomly initialized unlocked weights and no pre-training.
    \end{enumerate}
    The configuration above yielding the best results is then tested across different $\beta \in \{0.0, 0.1, 0.5, 1.0, 1.5, 3.0\}$ to observe the effects of scaling the KL divergence in pre-training of the feature extractor for the DRL agent. For the latter experiment, the parameters of the pre-trained encoder are locked. 
    
    Measuring the DRL agent's performance is done in three steps. Firstly, average episodic collision rates, cross-track errors, durations, and path progress during training over 3 million timesteps were compared. This corresponds to approximately 4,600-5,200 episodes, depending on the episodic durations of the specific agent. All agents are trained in a complex environment with 11 static and 17 dynamic obstacles placed along a curved path. The path goes through a random number of waypoints sampled from $\mathcal{U}(2,6)$, with the angle of the path at the starting point sampled from $\mathcal{U}(-\pi,\pi)$. Due to the complexity of the training environment, the convergence of the training error trajectories may be seen as an estimate of the actual performance of the agents. 
    
    Secondly, to evaluate their final performance, selected agents are tested in 100 randomly generated testing scenarios. Again, reporting statistics on path progress, cross-track error, duration, and collision rate. 
    
    Finally, to gain insights into the agents' behavior within the environment, their traversed trajectories are plotted in a select subset of the testing scenarios. These visualizations serve as a tool for qualitative evaluation, providing a more nuanced understanding of the agents' decision-making process, interaction dynamics, and consistency in the simulated environment.   
    
\section{Results and Discussion} \label{sec:Results}
    This section presents and discusses the obtained results and is divided into two main parts: VAE and DRL results. The former presents results from the VAE part of the work, and the latter underlines the results of the DRL agent when equipped with the VAE-based feature extractor.

    \subsection{VAE LiDAR reconstruction}        
\begin{figure}[h]
         \centering
         \begin{subfigure}{0.48\linewidth}
             \centering 
             \includegraphics[width=\linewidth]{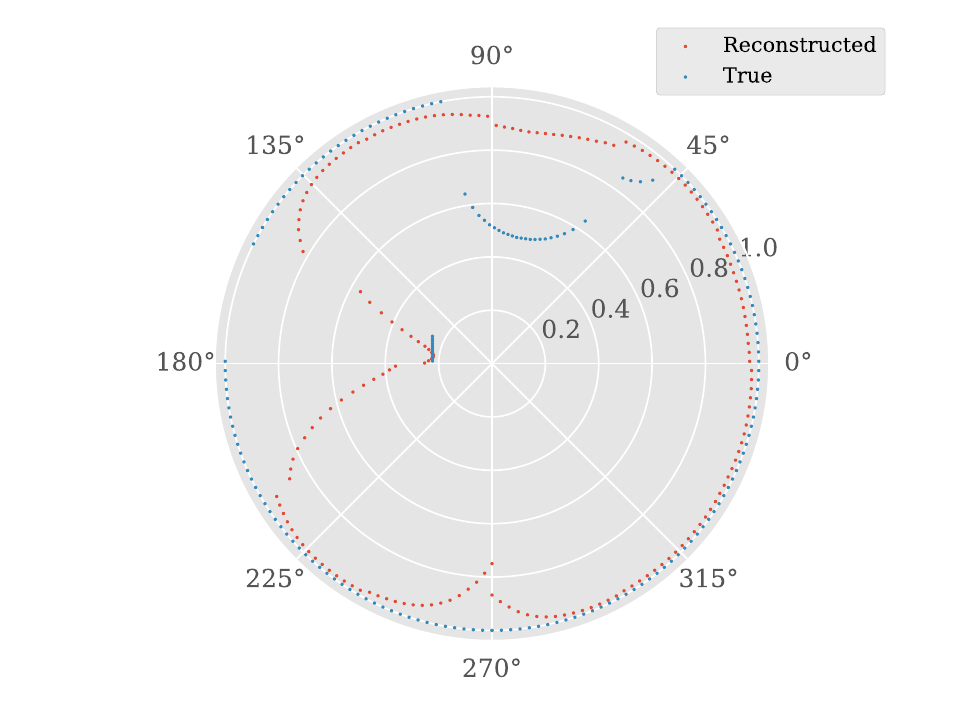}
             \caption{Zero-padding}
         \end{subfigure}
         \begin{subfigure}{0.48\linewidth}
             \centering
             \includegraphics[width=\linewidth]{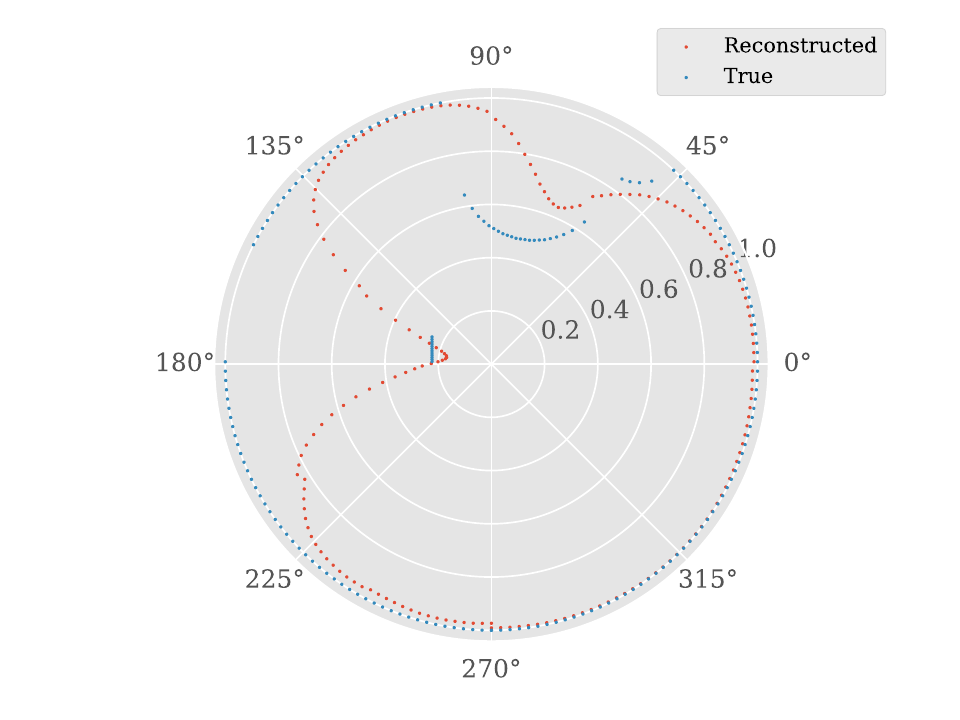}
             \caption{CPTC-1D}
         \end{subfigure}
        \caption{VAE reconstruction comparison between zero-padding and our adapted padding approach.}
        \label{fig:circular_pad_proof}
    \end{figure}

    Evaluating the effectiveness of the CPTC-1D method was done qualitatively by visually inspecting large numbers of reconstructions with and without the circularly padded decoder. \cref{fig:circular_pad_proof} displays a set of examples from this test, in which the inputs are in blue and reconstructions are in red. Two VAEs are presented: One using valid (zero-) padding in the decoder and one using CPTC-1D for the transposed convolutional layers of the decoder. 
    
    These results highlight the importance and efficiency of utilizing our adapted algorithm in the decoder for this specific data. \cref{fig:circular_pad_proof} highlights the edge artifacts produced with traditional zero-padding, causing discrepancies between inputs and reconstructions, particularly near the perception vector's start and endpoints. Applying the circular padding method significantly reduces these artifacts.

    \subsection{VAE training, validation, and testing}
    Moving forward with the CPTC-1D method, a set of VAEs is fitted to the training data as described in \cref{sec:Method:VAE_eval}. \cref{tab:test_error_basis} presents the mean errors and uncertainties for the VAEs.
    
    \begin{table}
        \centering
        \caption{VAE test errors across 10 random seeds for both models. The latent dimension is 12 and $\beta$ is set to 1.}
        \begin{tabular}{@{}lccc@{}}
        \toprule
         \textbf{Model (latent dim.)} & \textbf{Mean} & \textbf{95\% CI} & \textbf{Std. dev}\\
         \hline
         ShallowConvVAE (12)& 42.67 & [42.58, 42.75] & 0.097\\
         DeepConvVAE (12)& 45.13 & [40.72, 49.54] & 5.036\\
         \bottomrule
        \end{tabular}
        \label{tab:test_error_basis}
    \end{table}

    \subsection{$\beta-$regularization}
\begin{figure}
         \centering
         \begin{subfigure}[b]{0.3\linewidth}
             \centering
             \includegraphics[width=\linewidth]{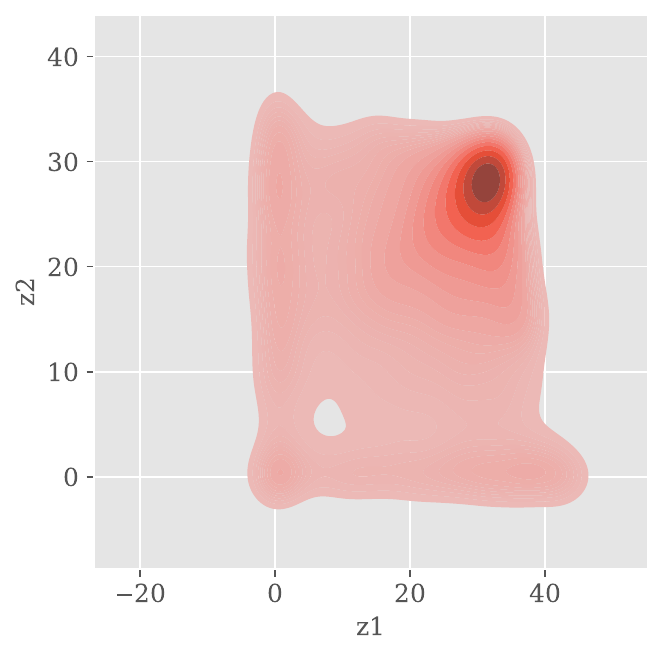}
             \caption{$\text{Shallow}, \beta = 0$}
         \end{subfigure}
         \hfill
         \begin{subfigure}[b]{0.3\linewidth}
             \centering
             \includegraphics[width=\linewidth]{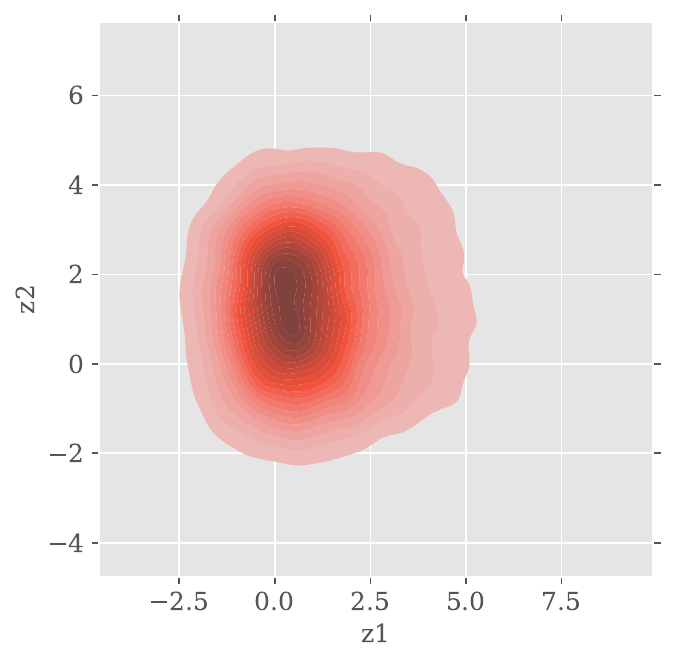}
             \caption{$\text{Shallow}, \beta = 1$}
         \end{subfigure}
         \hfill
         \begin{subfigure}[b]{0.3\linewidth}
             \centering
             \includegraphics[width=\linewidth]{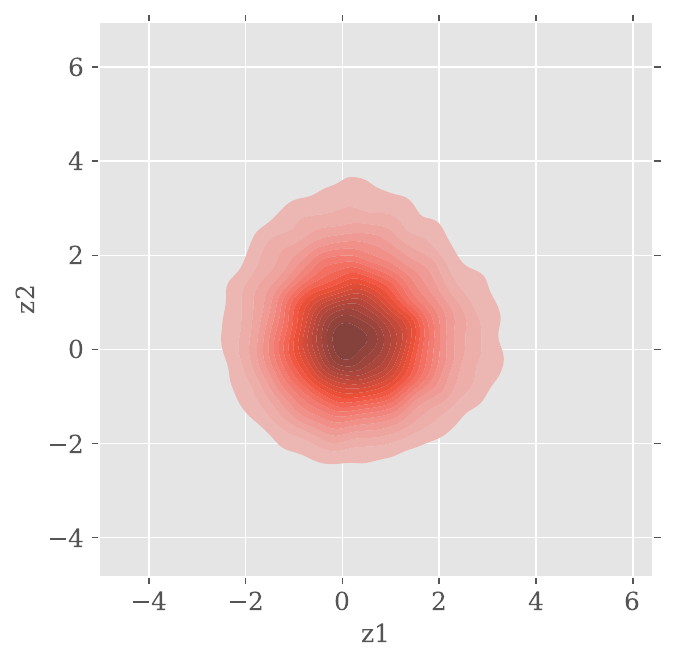}
             \caption{$\text{Shallow}, \beta = 5$}
         \end{subfigure}
    
         \begin{subfigure}[b]{0.3\linewidth}
             \centering
             \includegraphics[width=\linewidth]{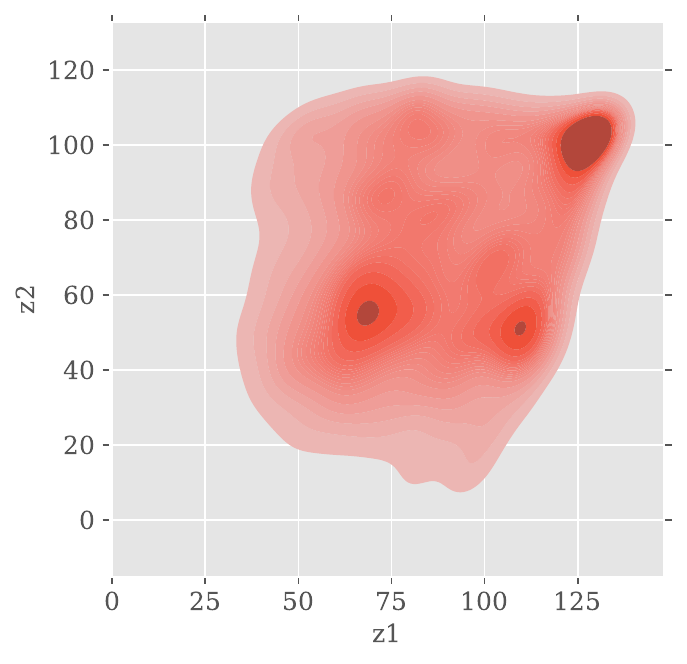}
             \caption{$\text{Deep}, \beta = 0$}
         \end{subfigure}
         \hfill
         \begin{subfigure}[b]{0.3\linewidth}
             \centering
             \includegraphics[width=\linewidth]{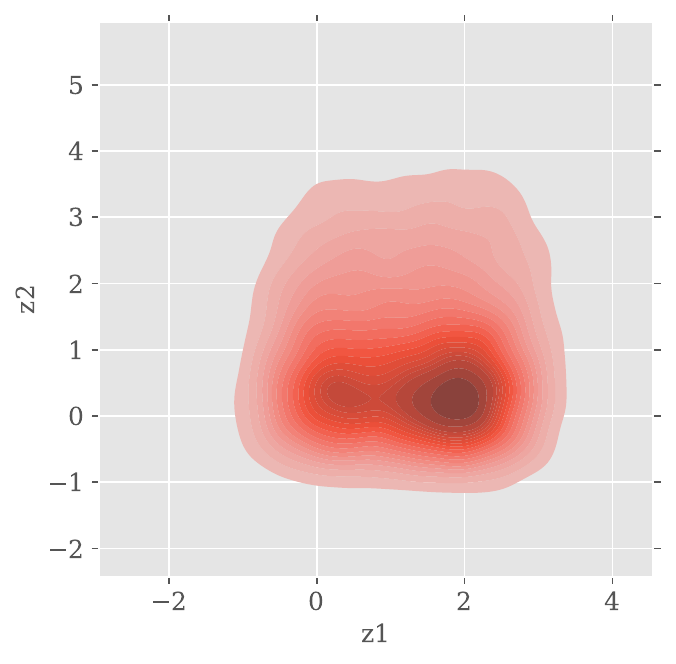}
             \caption{$\text{Deep}, \beta = 1$}
         \end{subfigure}
         \hfill
         \begin{subfigure}[b]{0.3\linewidth}
             \centering
             \includegraphics[width=\linewidth]{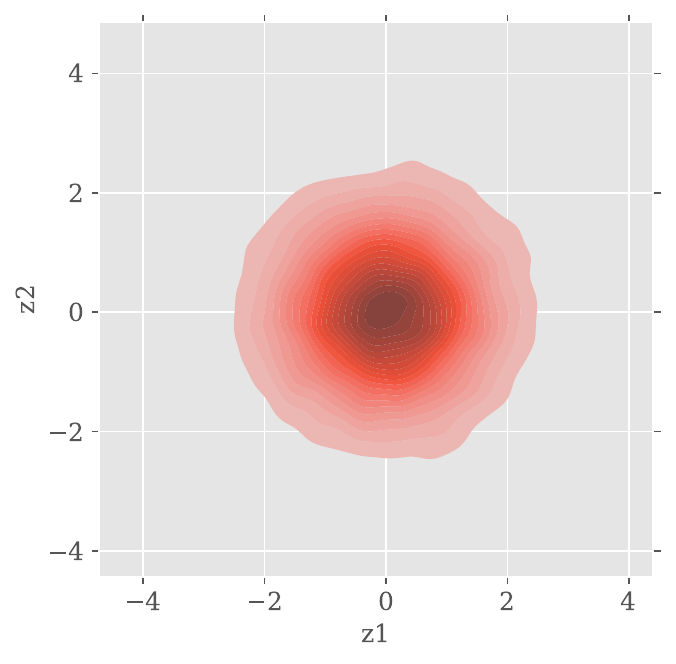}
             \caption{$\text{Deep}, \beta = 5$}
         \end{subfigure}
            \caption{Kernel density estimates of the latent representation of the test-set for six distinct VAEs.}
            \label{fig:betas_latent_space}
    \end{figure}

    To assess the impact of $\beta$-regularization on the latent space encoding, a set of VAEs is trained across a range of $\beta$ values. Visualizing high-dimensional latent spaces quickly becomes impractical as the dimensionality increases; to showcase the effect of $\beta$-regularization, the VAEs in this demonstration is constrained to learn only two latent features. \cref{fig:betas_latent_space} shows the estimated latent distributions for the test set across six distinct VAEs trained for 25 epochs on the dataset, with $\beta \in \{0, 1, 5\}$. This figure's top row shows results from the shallow model with increasing $\beta$, while the bottom row depicts results from the deep model.
    
    When $\beta = 0$, the VAE behaves as a standard autoencoder without regularization. As $\beta$ increases, the kernel density estimate of the latent representations from the test set progressively converges towards an isotropic multivariate Gaussian distribution, implying a shift in the latent distribution itself. In the presented 2D context, opting for a $\beta$ in the proximity of 1 appears to strike a balance, preserving information about specific inputs while at the same time preventing overfitting. Note that these results are specific to a two-dimensional latent space, a configuration chosen solely for visualization. The latent space will take on other shapes as the latent dimensionality increases. Also, note that for very small $\beta \approx 0$, where the VAE approximates an autoencoder, the narrow bottleneck acts as an implicit regularizer itself, suggesting that higher dimensional latent distributions may have more complex (non-convex) shapes. Nonetheless, these results still provide valuable insights into the latent representations specific to our dataset.

    \subsection{Avoiding posterior collapse}
\begin{figure}[h]
        \centering
        \begin{subfigure}[b]{\linewidth}
            \centering
            \includegraphics[width=\linewidth]{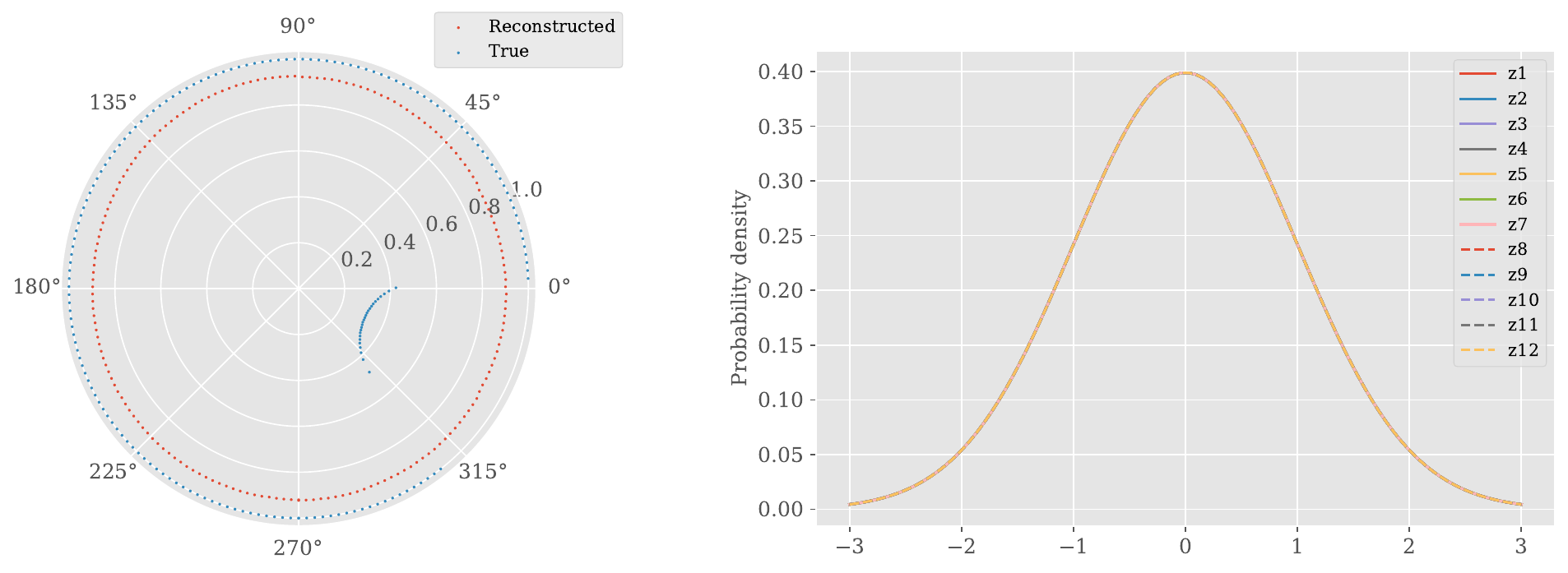}
            \caption{Collapsed latent features}
            \label{subfig:mode_collapse_example1}
        \end{subfigure}
    
        \begin{subfigure}[b]{\linewidth}
            \centering
            \includegraphics[width=\linewidth]{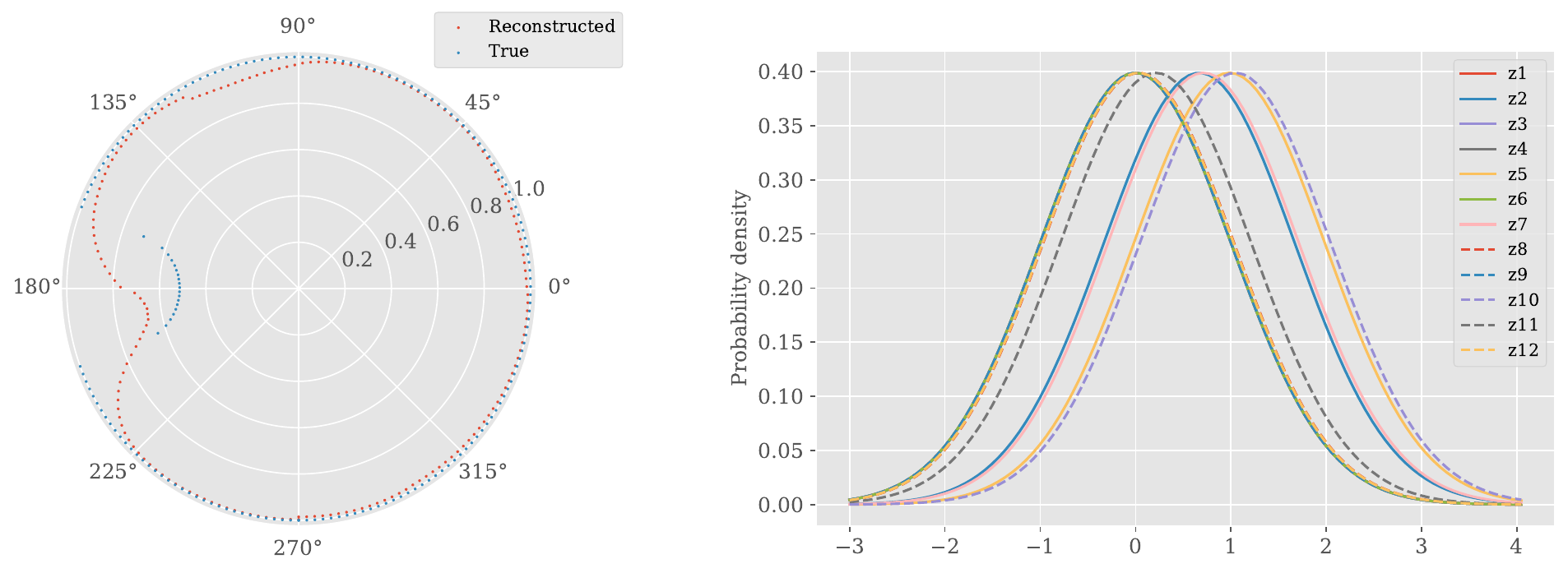}
            \caption{Expected behavior}
            \label{subfig:mode_collapse_example2}
        \end{subfigure}
        \caption{Posterior collapse}
        \label{fig:mode_collapse_example}
    \end{figure}

    In preliminary investigations, some VAE models suffered from near-\textit{posterior collapse} \cite{wang2021posteriorCollapse}. Posterior collapse describes the phenomenon where all observations are mapped to identical latent distribution, namely the prior ($\mathcal{N}(\boldsymbol{0}, \boldsymbol{I})$). This collapse renders the latent encoding essentially void of information pertaining to the input data, as illustrated in \cref{subfig:mode_collapse_example1}.
    Avoiding this collapse is a manner of choosing the appropriate $\beta$. \cref{subfig:mode_collapse_example2} shows the expected behavior characterized by non-collapsing latent distributions generated by the encoder, as the latent distributions vary considerably between distinct inputs. This sensitivity implies that the latent representations carry input-specific information, thereby making them \textit{informative} in this context.
    
    \subsection{Deep Reinforcement Learning}
    This section presents the results from training and testing the DRL agent equipped with several feature extractors from the preceding VAE section.
    
    \subsubsection{Training metrics}
\begin{figure}
         \centering
         \begin{subfigure}[b]{0.48\linewidth}
             \centering
             \includegraphics[width=\linewidth]{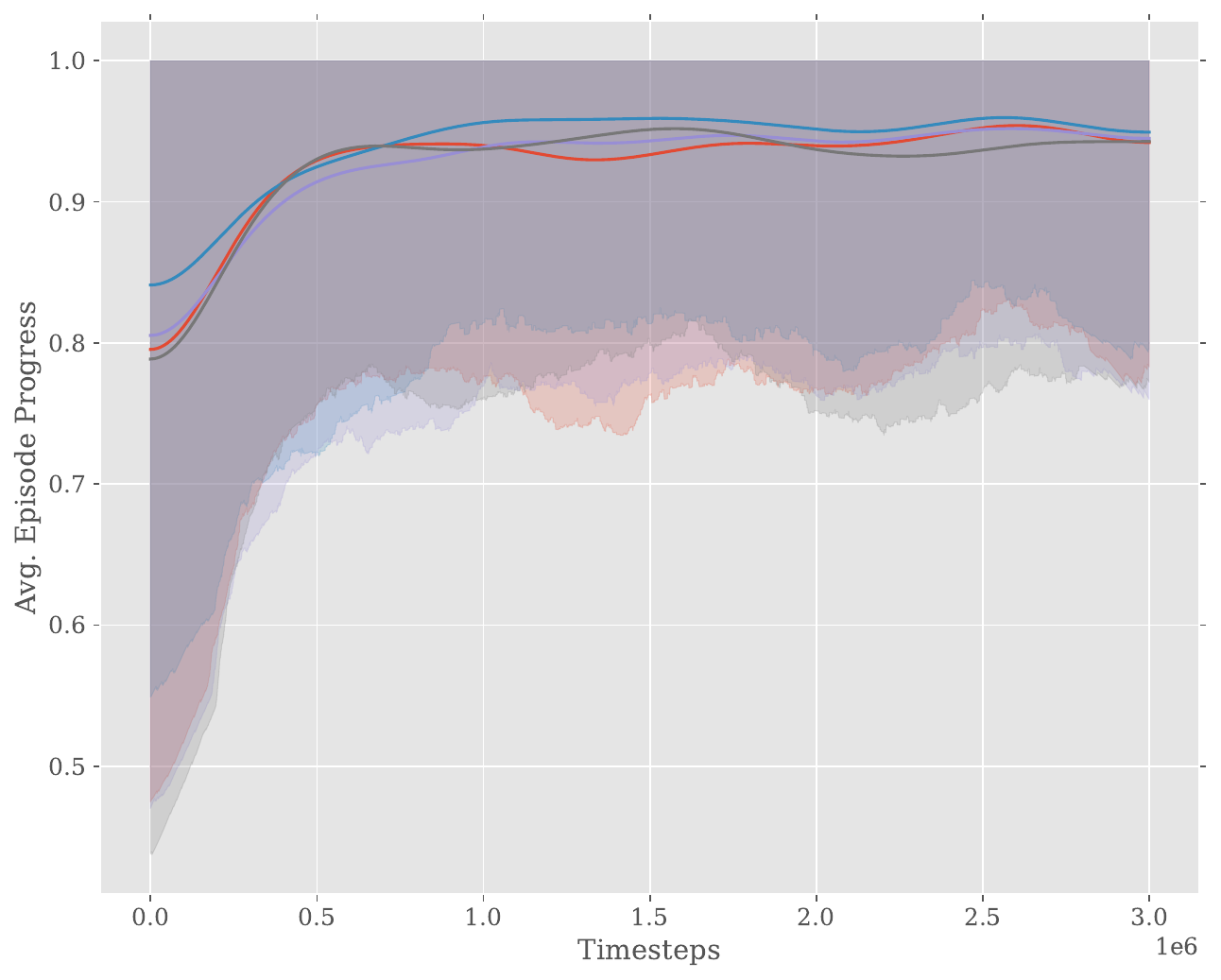}
             \caption{Path progress}
         \end{subfigure}
         \hfill
         \begin{subfigure}[b]{0.48\linewidth}
             \centering
             \includegraphics[width=\linewidth]{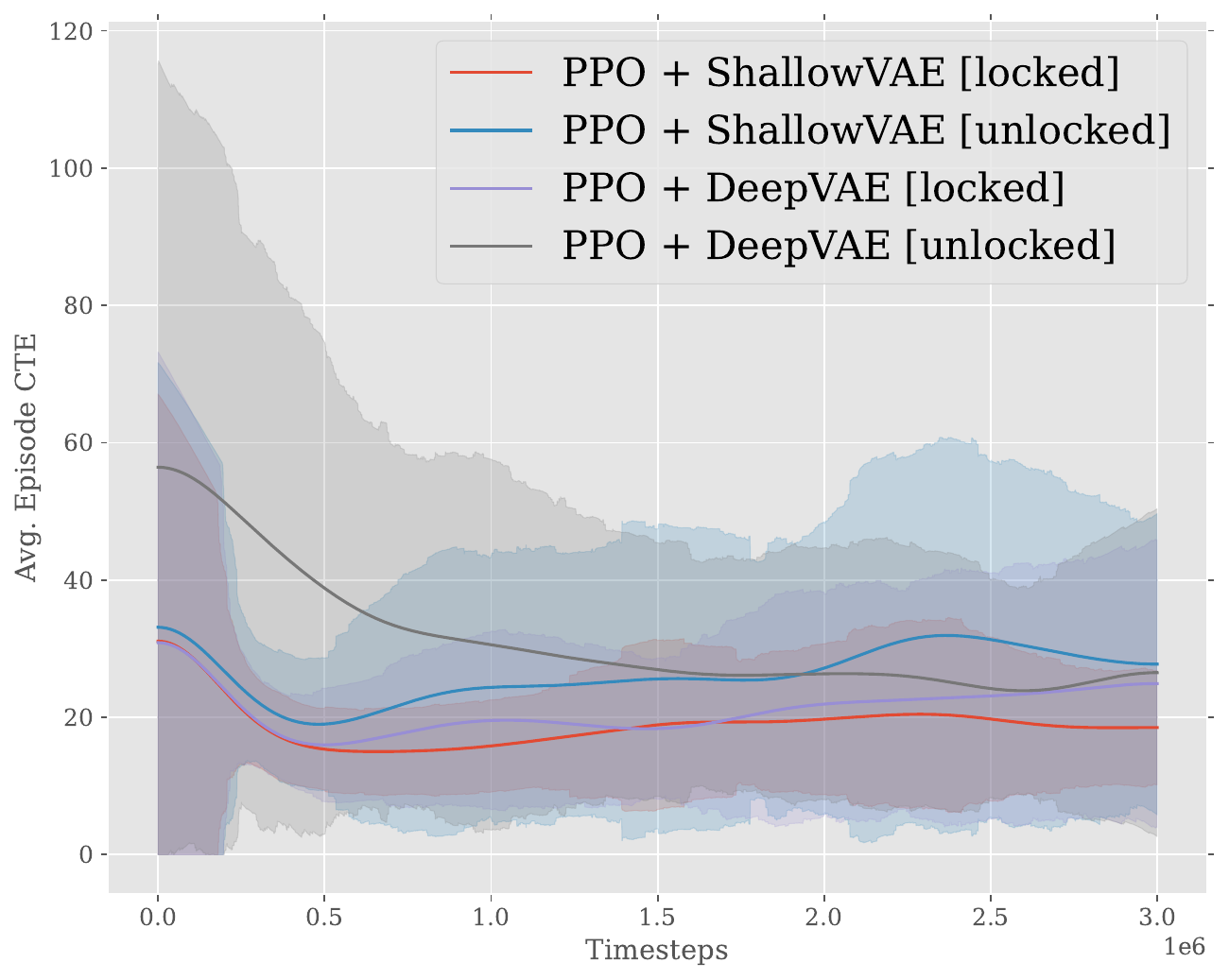}
             \caption{CTE}
         \end{subfigure}
         
         \begin{subfigure}[b]{0.48\linewidth}
             \centering
             \includegraphics[width=\linewidth]{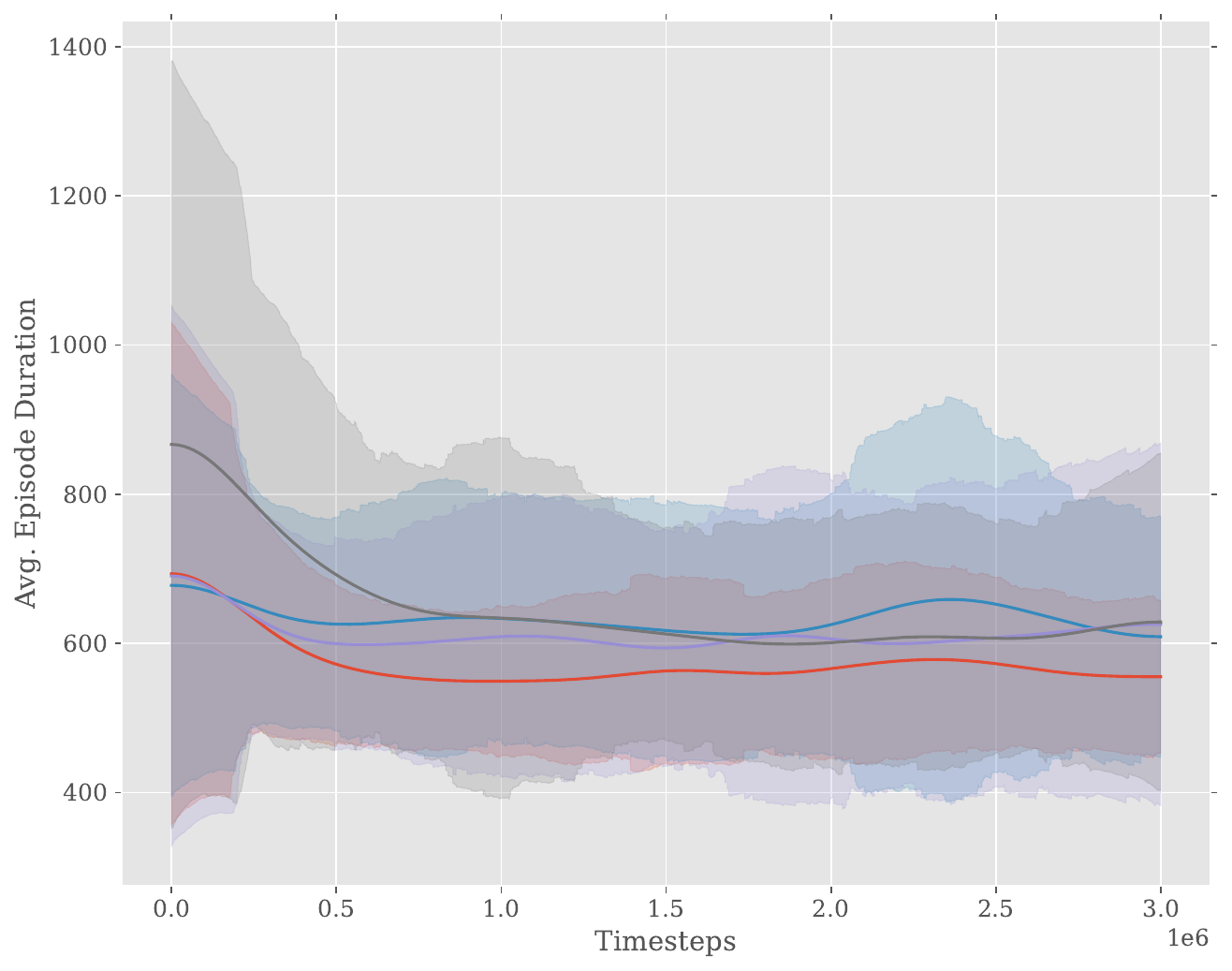}
             \caption{Duration}
         \end{subfigure}
         \hfill
         \begin{subfigure}[b]{0.48\linewidth}
             \centering
             \includegraphics[width=\linewidth]{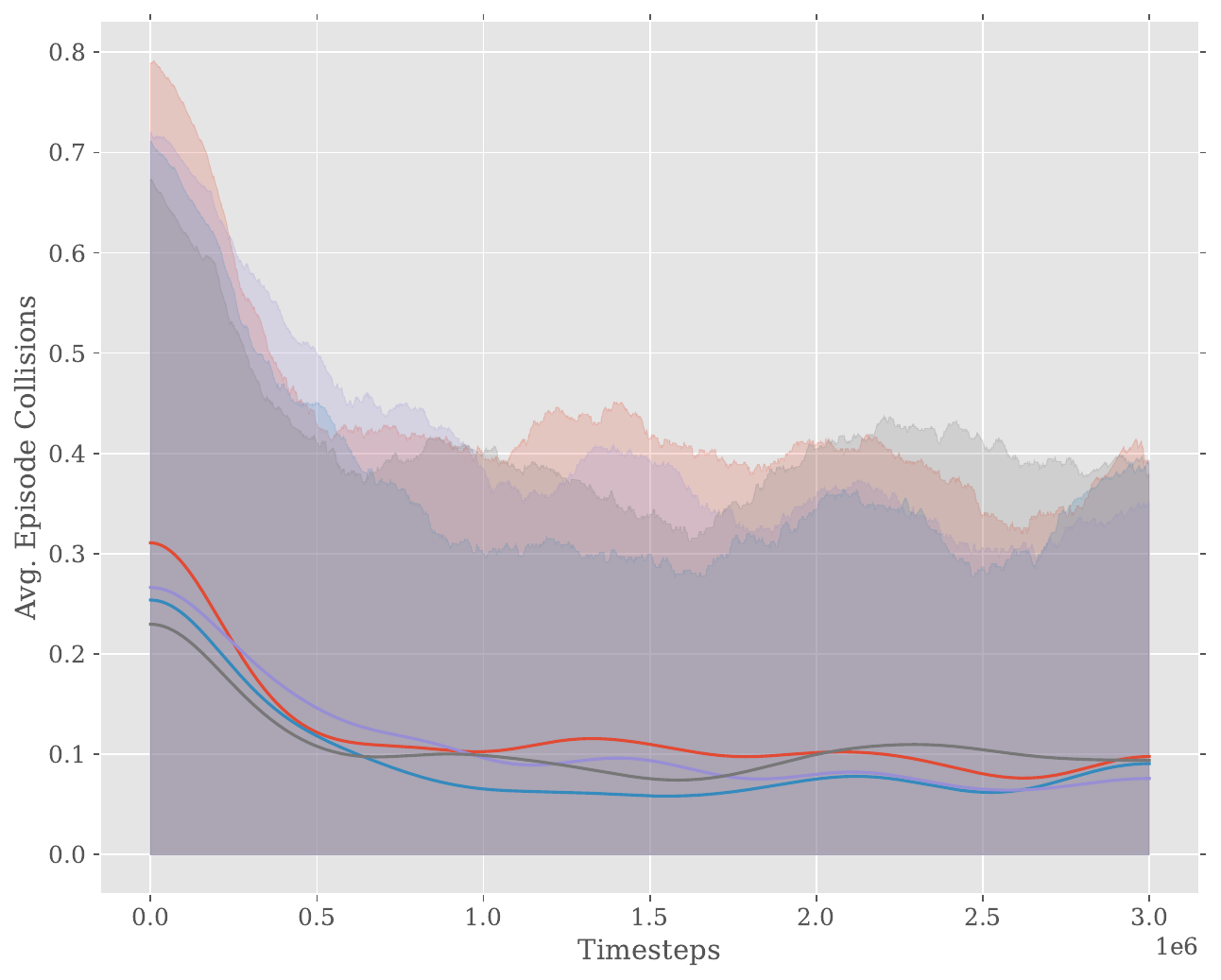}
             \caption{Collision rate}
         \end{subfigure}
    
            \caption{Training evolutions of the performance metrics for the four different DRL configurations.}
            \label{fig:4_config_drl_results}
    \end{figure}

    Four configurations are compared: Shallow locked, Shallow unlocked, Deep locked, and Deep unlocked, as described in \cref{sec:performance_metrics_drl}. The tests resulted in the performance metrics plotted in \cref{fig:4_config_drl_results}. The metrics are plotted as Gaussian smoothed rolling averages over 100 episodes, with associated rolling standard deviations represented by the shaded areas.
    
    The models perform similarly in the path progress metric; all reporting averages around $95\%$ with similar standard deviations. For CTE, the two locked models seem to perform better than the unlocked ones. The gradual increase in CTE with increasing episodes suggests that the agent learns to deviate from the path to gain a higher total reward. For the average episodic duration, the shallow locked model outperforms the three others. Collision rates do not differ significantly across models. A likely explanation for the slight performance difference between locked and unlocked configurations is that locking the feature extractor's parameters substantially reduces the searchable parameter space. This reduction allows the agent to more effectively explore a larger portion of the parameter space. With locked feature extractor parameters, the agent can focus on finding a good policy rather than optimizing the feature extractor. 

    Although the initial rationale for unlocking the (pre-trained) parameters was to give the agent a head start in identifying optimal feature extractor parameters, our findings suggest that this approach does not confer a significant advantage in the given context. Considering that the VAE pre-training is designed to obtain an effective encoding, the agent's need to further adjust this encoding might be redundant. This assertion is supported by the results from the unlocked training, where no significant advantages were observed.

    \subsubsection{VAE encoder vs. baseline PPO agent}
\begin{figure}[h]
         \centering
         \begin{subfigure}[b]{0.48\linewidth}
             \centering
             \includegraphics[width=\linewidth]{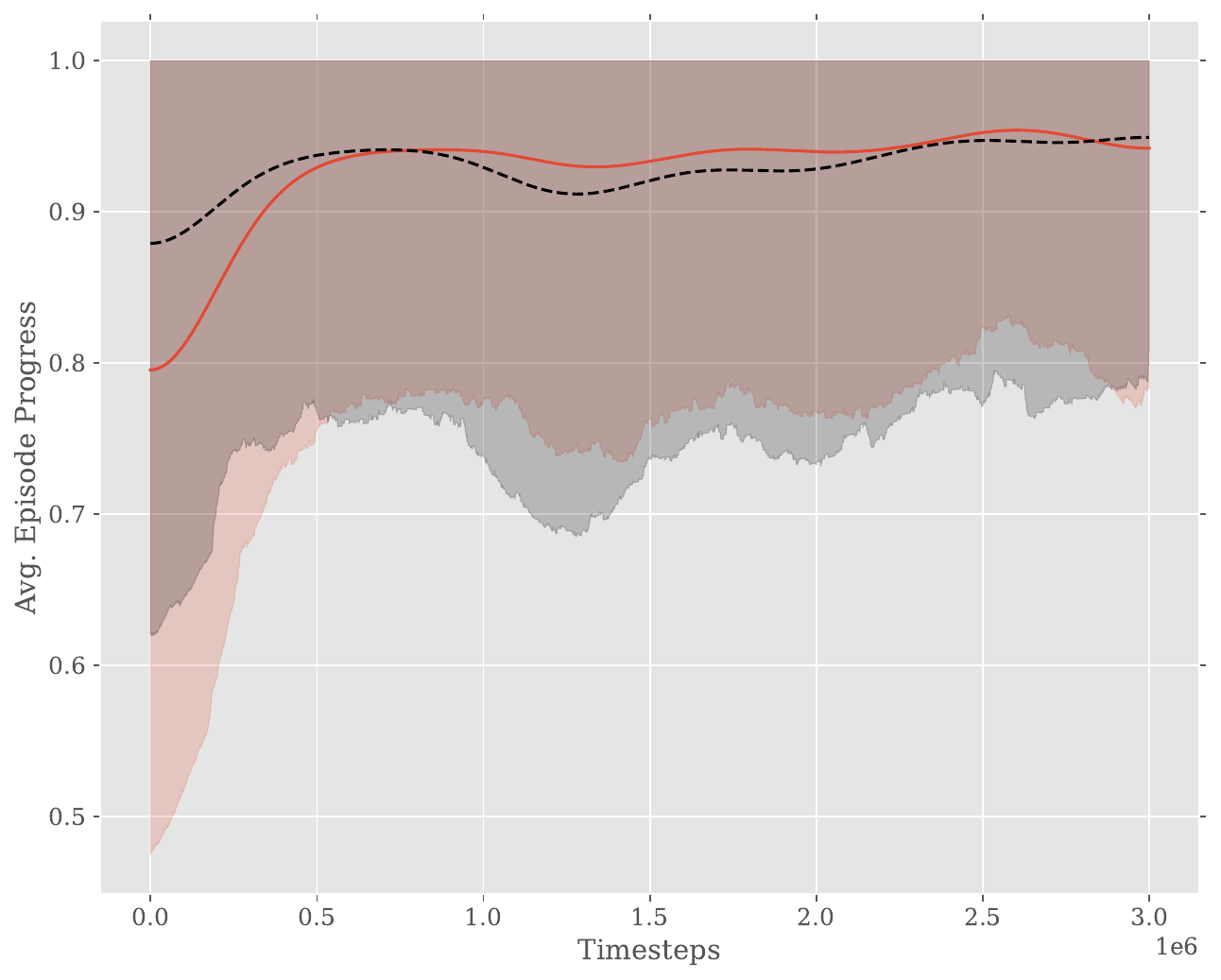}
             \caption{Path progress}
             \label{subfig:progress_rate_drl_baseline_vs_ours}
         \end{subfigure}
         \hfill
         \begin{subfigure}[b]{0.48\linewidth}
             \centering
             \includegraphics[width=\linewidth]{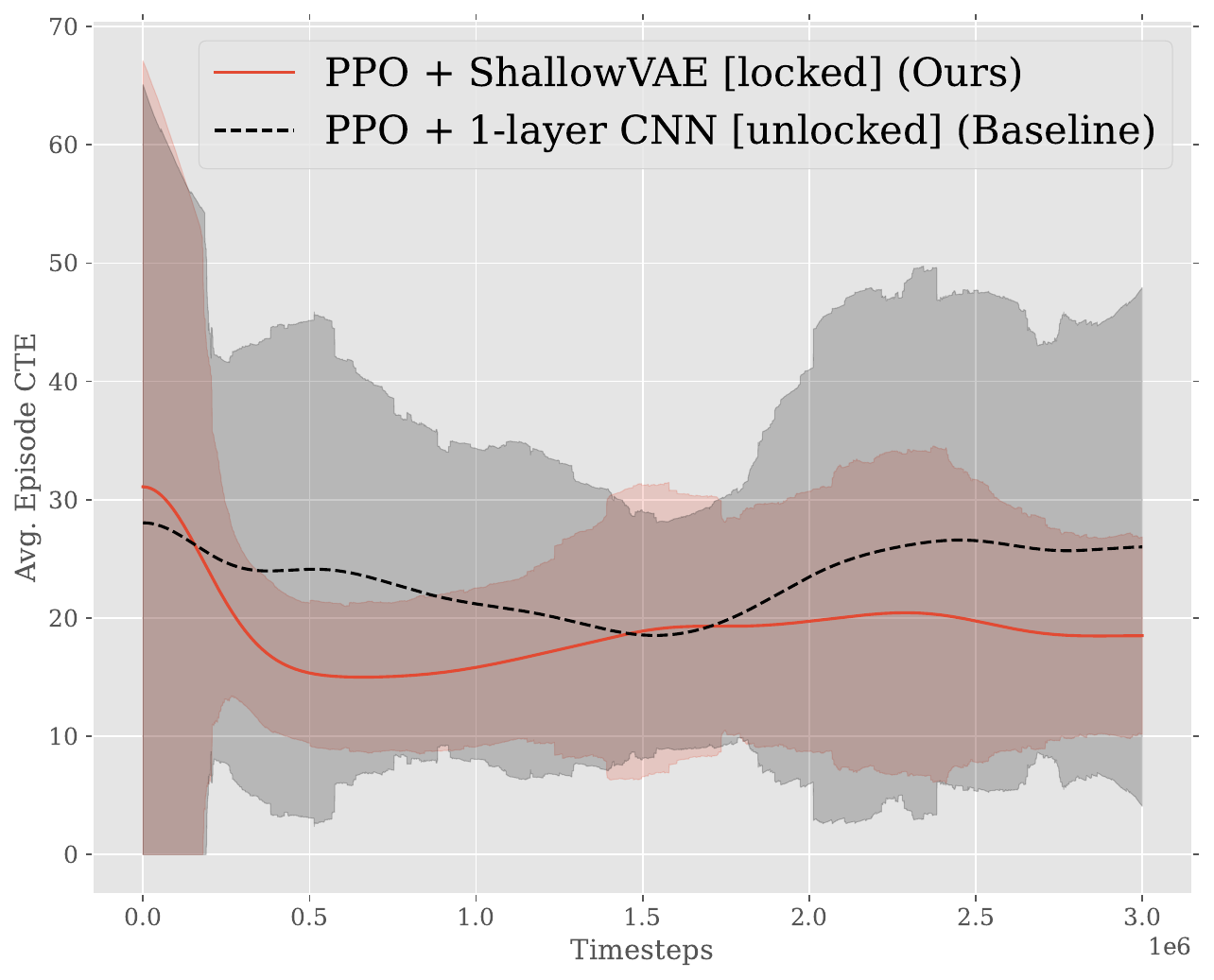}
             \caption{CTE}
         \end{subfigure}
         
         \begin{subfigure}[b]{0.48\linewidth}
             \centering
             \includegraphics[width=\linewidth]{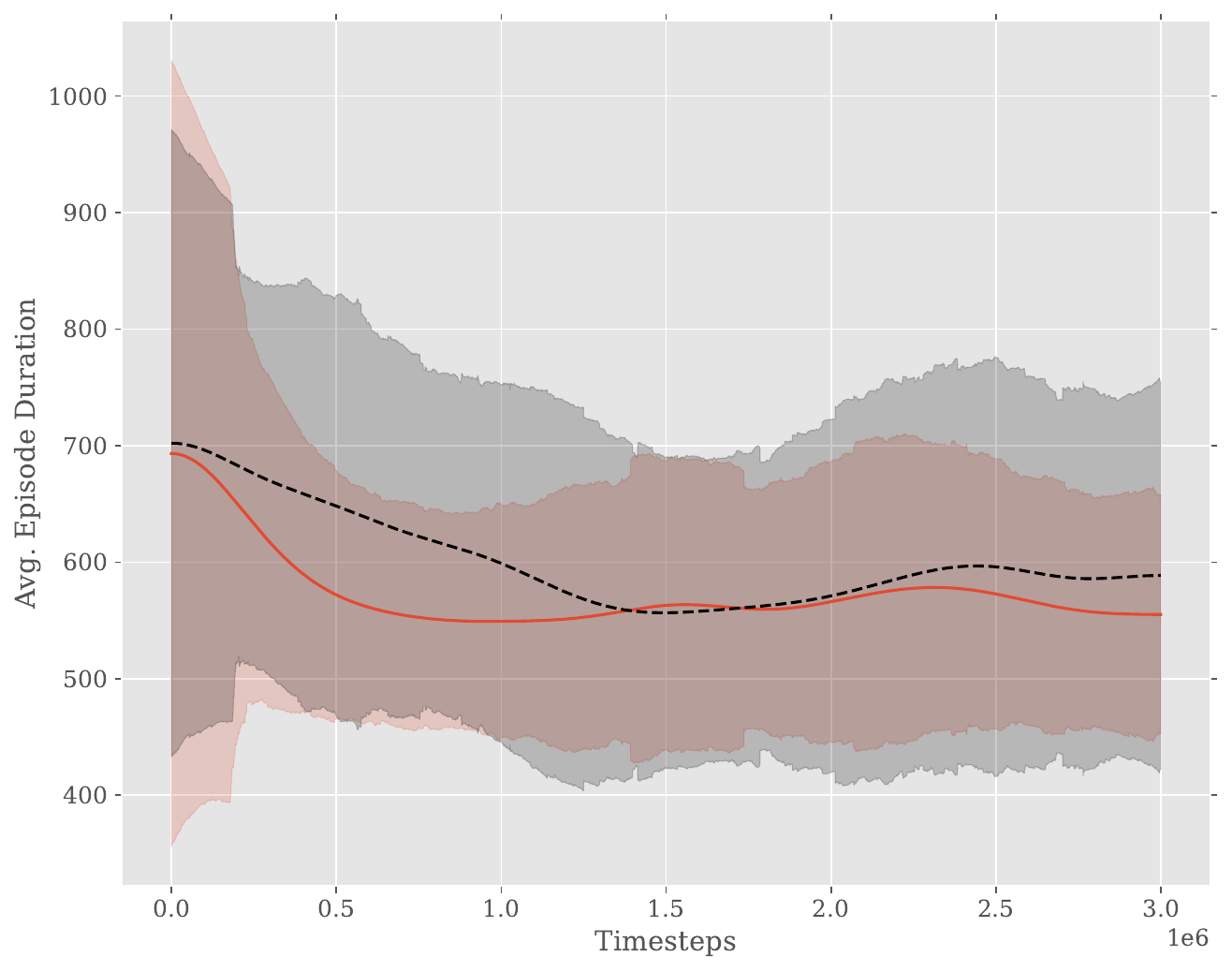}
             \caption{Duration}
         \end{subfigure}
         \hfill
         \begin{subfigure}[b]{0.48\linewidth}
             \centering
             \includegraphics[width=\linewidth]{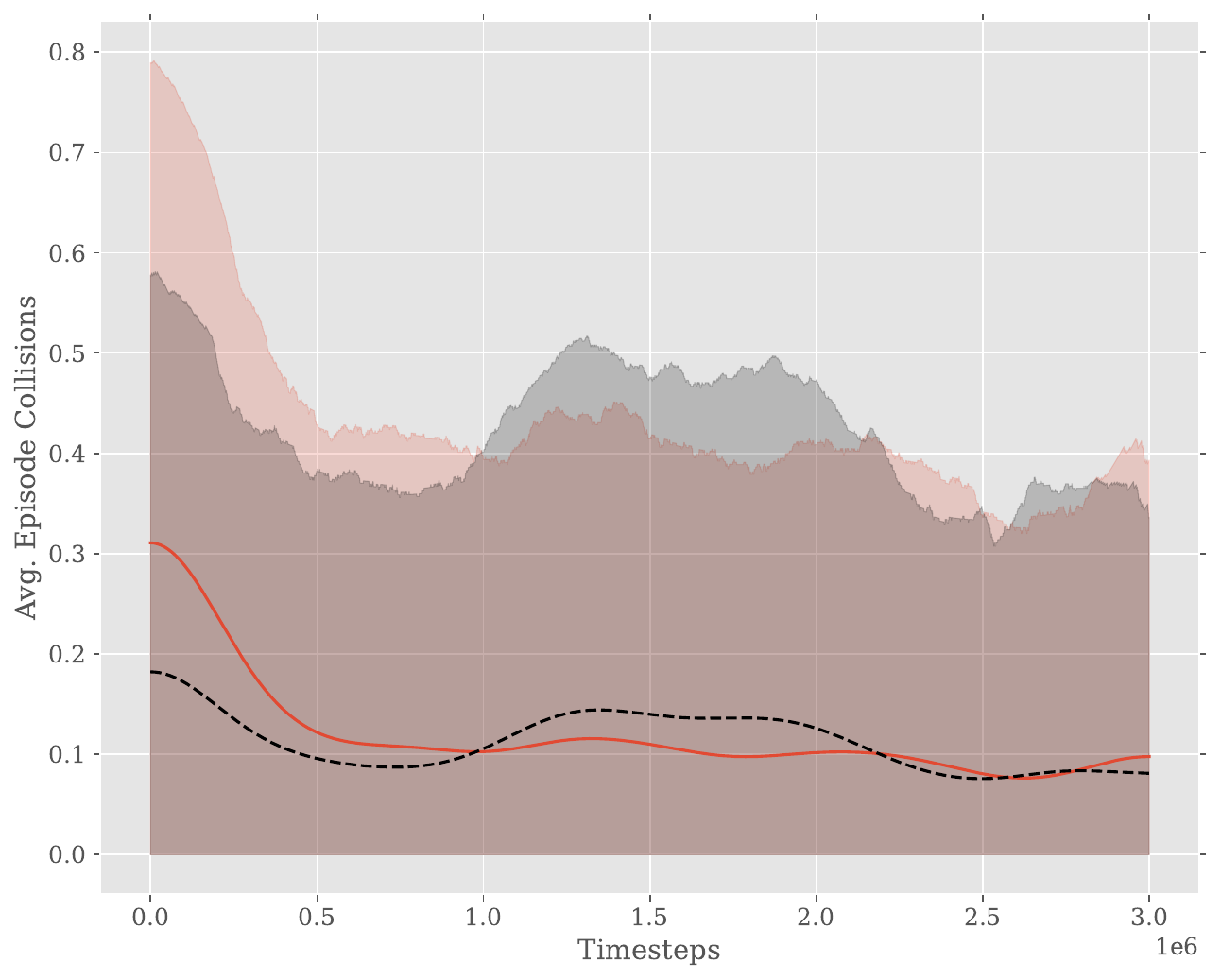}
             \caption{Collision rate}
             \label{subfig:collision_rate_drl_baseline_vs_ours}
         \end{subfigure}
    
            \caption{Training evolutions of the performance metrics for our best-performing model and the baseline.}
            \label{fig:drl_result_ours_vs_baseline}
    \end{figure}

    The metrics of our best-performing configuration, ShallowConvVAE [locked], are plotted against the baseline in \cref{fig:drl_result_ours_vs_baseline} to estimate the potential advantage gain of the VAE approach.
    
    The plot reveals a performance difference in CTE and duration, reporting consistently lower averages and variances for our model. Based on these observations, it is reasonable to infer that our model yields better path adherence and more efficient trajectories than the baseline. Low CTE and duration values might indicate a tendency toward high-risk behavior by the agent, potentially leading to safety concerns. However, \cref{subfig:progress_rate_drl_baseline_vs_ours} and \cref{subfig:collision_rate_drl_baseline_vs_ours} show that our model exhibits a similar path progress and collision rate compared to the baseline. This observation supports the notion that our model achieves suitable trajectories without compromising safety compared to the baseline. To fully understand the risk vs. efficiency trade-off presented by these findings, a qualitative assessment of the actual paths in the environment is needed.

    \subsubsection{DRL performance evaluation}
\begin{figure}[h!]
         \centering
    
         \begin{subfigure}[b]{0.49\linewidth}
             \centering
             \includegraphics[width=\linewidth]{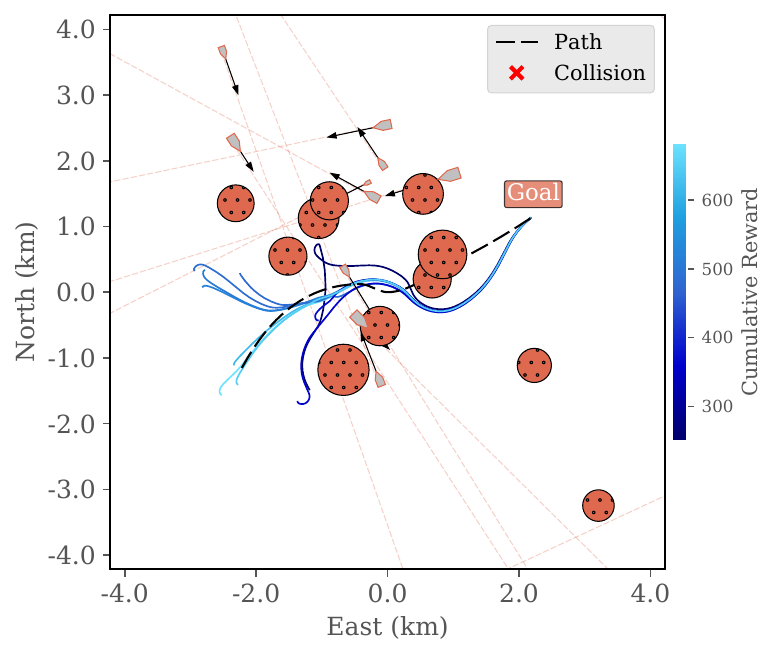}
             \caption{Shallow [locked]}
         \end{subfigure}
         \hfill
         \begin{subfigure}[b]{0.49\linewidth}
             \centering
             \includegraphics[width=\linewidth]{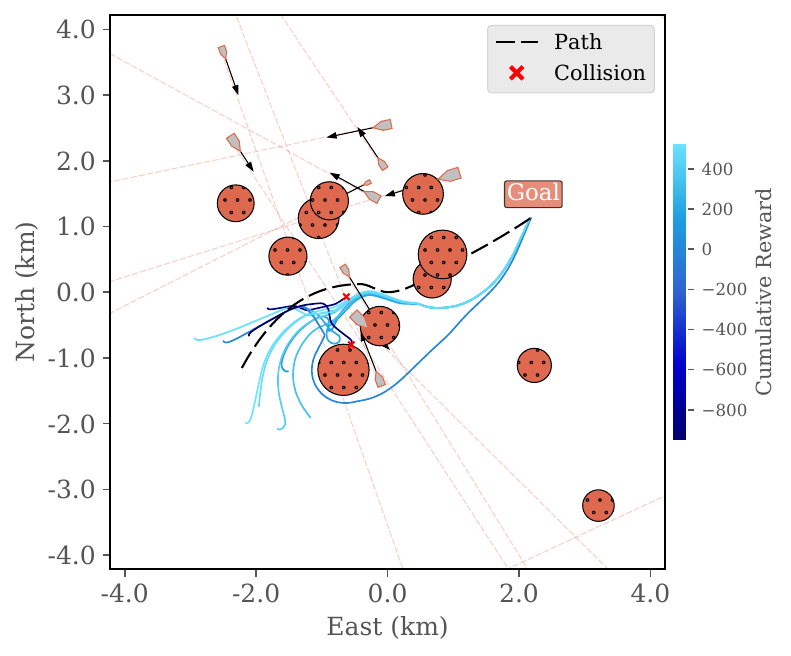}
             \caption{Deep [locked]}
         \end{subfigure}
    
         \begin{subfigure}[b]{0.49\linewidth}
             \centering
             \includegraphics[width=\linewidth]{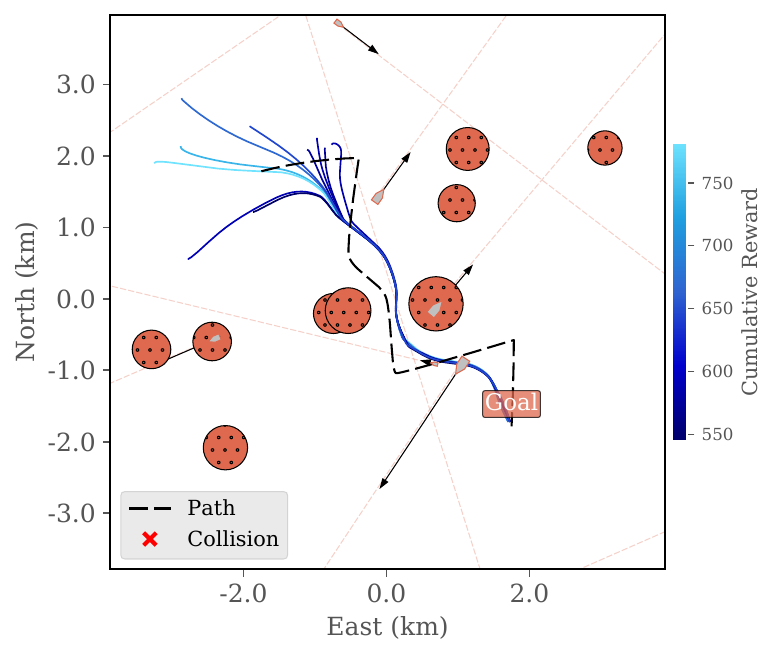}
             \caption{Shallow [locked]}
         \end{subfigure}
         \hfill
         \begin{subfigure}[b]{0.49\linewidth}
             \centering
             \includegraphics[width=\linewidth]{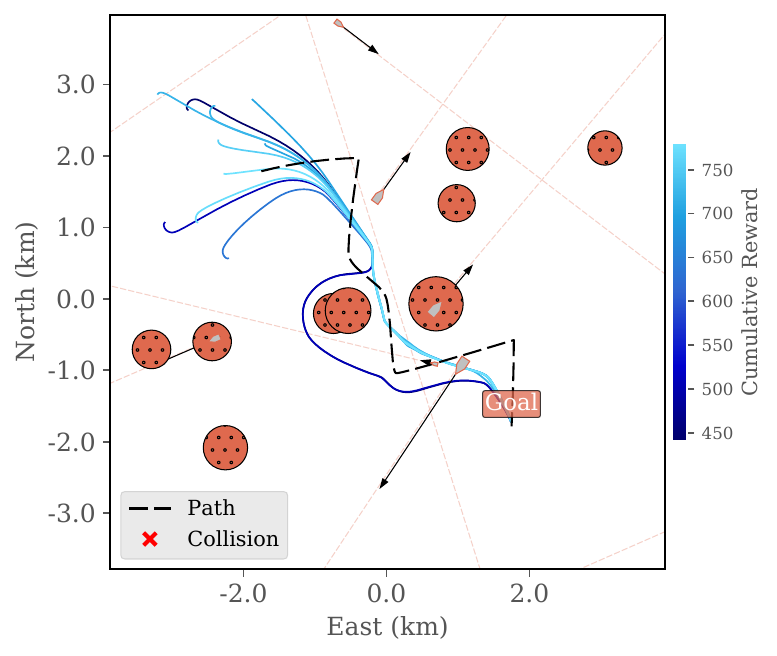}
             \caption{Deep [locked]}
         \end{subfigure}
         
         \begin{subfigure}[b]{0.49\linewidth}
             \centering
             \includegraphics[width=\linewidth]{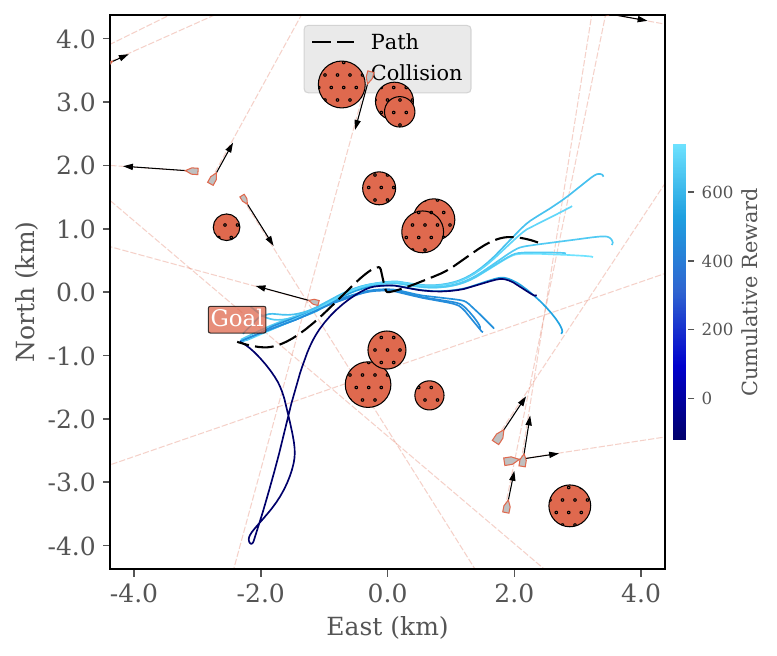}
             \caption{Shallow [locked]}
             \label{subfig:example_trajs_sl3}
         \end{subfigure}
         \hfill
         \begin{subfigure}[b]{0.49\linewidth}
             \centering
             \includegraphics[width=\linewidth]{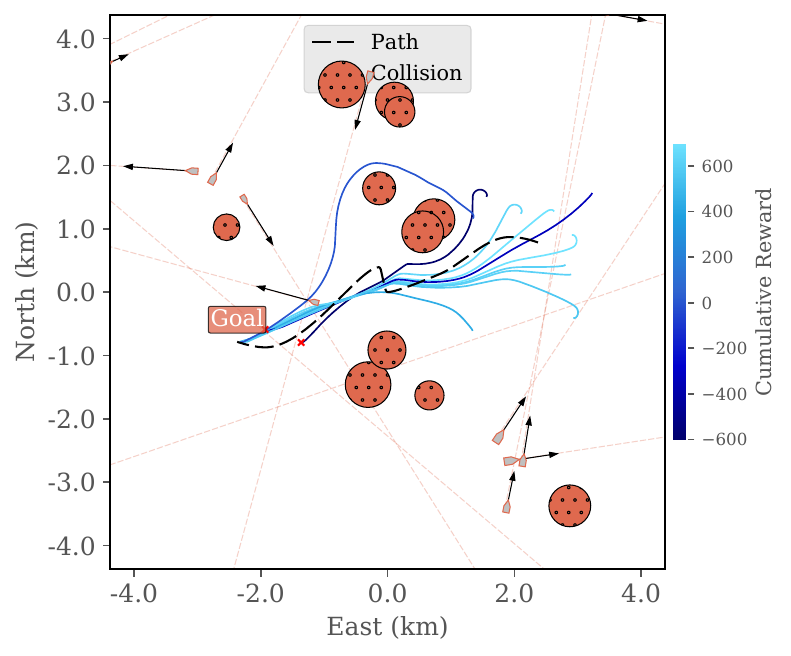}
             \caption{Deep [locked]}
         \end{subfigure}
        
        \caption{Trajectory comparisons between encoder architectures with randomized vessel initial states.}
        \label{fig:example_trajs}
    \end{figure}

    First, to showcase the consistency in trajectories between the proposed models, \cref{fig:example_trajs} shows trajectories for a set of random environments and perturbed initial states of the ASV. Each trajectory in a given model-environment combination (i.e., each subplot) is colored by the cumulative reward the specific agent received. These results show how the deep configuration is more prone to larger deviations from the path to circumvent obstacles compared to the shallow model, while there still is some unintended behavior (\cref{subfig:example_trajs_sl3}) occurring for the shallow model, too.
    
    Lastly, the test performance of the following models: ShallowConvVAE + PPO [locked], DeepConvVAE + PPO [locked], and the baseline (1-layer CNN [unlocked], no pre-training) is quantitatively compared. The test results are given in \cref{tab:drl_testresults}, reported as averages over 100 random scenarios. These results are similar to the convergence of the training trajectories, and the shallow model is, again, excelling in path adherence and reports low average episodic durations. However, the shallow model recorded slightly worse collision avoidance capabilities than the baseline. This suggests a potential trade-off between efficient trajectories and collision risk. Despite this, the shallow model showed a marginal improvement over the baseline in path progress.

    \begin{table*}[h!]
        \centering
        \caption{Test results from 100 episodes, reported as means with 95\% confidence intervals in the square brackets.}
        \begin{tabular}{lllllll}
        \toprule
        \textbf{Metric} & \multicolumn{2}{l}{\textbf{ShallowConvVAE}}& \multicolumn{2}{l}{\textbf{DeepConvVAE}} & \multicolumn{2}{l}{\textbf{Baseline}}\\
        \hline
        Progress & $\mathbf{96.7}$\textbf{\%} & $\mathbf{[94.5, 98.9]}$ & $94.4 \%$ & $[90.8, 97.9]$ & $96.4\%$ & $[93.8, 99.1]$ \\
        CTE & $\mathbf{18.7}$ & $\mathbf{[16.6, 20.9]}$ & $22.7$ & $[18.6, 26.8]$ & $21.4$ & $[18.9, 24.0]$\\
        Duration & $\mathbf{547.2}$ & $\mathbf{[530.4, 563.9]}$ & $557.5$ & $[526.2, 588.8]$ & $557.6$ & $[539.9, 575.2]$\\
        Collisions & $5.0$\% & $[0.7, 9.4]$ & $7.0\%$ & $[1.9, 12.1]$ & $\mathbf{4.0}\textbf{\%}$ & $\mathbf{[0.1, 7.9]}$\\
        \bottomrule
        \end{tabular}
        \label{tab:drl_testresults}
    \end{table*}

\section{Conclusion} \label{sec:Conclusion}
The main conclusions of this study in light of the set research questions can be itemized as follows:
\begin{itemize}
    \item \textit{Can the VAE feature extractor be successfully integrated into a DRL agent within the simulated maritime environment?} The study successfully addressed this question by integrating a pre-trained VAE feature extractor into the DRL framework, demonstrating its effectiveness in achieving efficient path adherence and collision avoidance in the simulated maritime environment.
    \item \textit{How do model complexity and hyperparameters influence the VAE's data reproduction and generalization capabilities?} The study explored the influence of model complexity and hyperparameters on the VAE's data reproduction and generalization capabilities. Circularly padded transposed convolution improved data reconstruction, and both shallow and deep VAE models exhibited good generalization abilities. The selection of an appropriate $\beta$ value was highlighted as crucial to prevent posterior collapse and maintain informative latent representations. 
    \item \textit{What is the performance difference between DRL agents equipped with pre-trained VAE-encoders of different complexity, and how do they compare to a non-VAE DRL agent?} Comparative analysis revealed that the PPO + ShallowConvVAE configuration with locked parameters was the most effective, outperforming both its unlocked counterpart and the non-VAE DRL agent in path adherence and efficiency. However, a slight increase in collision rate during testing was observed, indicating a potential inclination towards more risk-taking behaviors that warrant further investigation.
\end{itemize}

In summary, the work effectively addresses all three research questions, providing insights into the successful integration of VAE with DRL, the impact of model complexity and hyperparameters on VAE capabilities, and the performance differences between DRL agents with pre-trained VAE-encoders vs. a non-VAE DRL agent.

\section*{Acknowledgments}
This work is part of SFI AutoShip, an 8-year research-based innovation center. The authors thank the partners, including the Research Council of Norway, under project number 309230.


\bibliographystyle{asmeconf}
\end{document}